\documentclass[11pt]{article}

\usepackage{color}
\usepackage[round]{natbib}
\usepackage[margin=1.1in]{geometry}

\usepackage{booktabs} 
\usepackage{subcaption}
\usepackage[table]{xcolor}
\usepackage{amsmath, amsfonts, bbm}
\usepackage{placeins}
\usepackage{hyperref}

\usepackage{float}
\usepackage{etoolbox, xspace}
\usepackage[textsize=scriptsize,textwidth=2.2cm]{todonotes}
\newtoggle{showtodos}
\toggletrue{showtodos}
\usepackage{makecell}
\usepackage{etoc}
\usepackage{longtable}

\iftoggle{showtodos}{
    \newcommand{\ludwig}[1]{\todo[color=blue!40]{Ludwig: #1}}
    \newcommand{\john}[1]{\todo[color=orange!40]{John: #1}}
    \newcommand{\karl}[1]{\todo[color=yellow!40]{Karl: #1}}
}{
    \newcommand{\ludwig}[1]{}
    \newcommand{\john}[1]{}
    \newcommand{\karl}[1]{}
}

\newcommand{\squad}{SQuAD\xspace}
\newcommand{\dist}{\mathcal{D}}
\newcommand{\nytdrop}{3.8\xspace}
\newcommand{\redditdrop}{14.0\xspace}
\newcommand{\amazondrop}{17.4\xspace}
\newcommand{\wikislope}{0.92\xspace}
\newcommand{\nytslope}{1.02\xspace}
\newcommand{\redditslope}{1.19\xspace}
\newcommand{\amazonslope}{1.36\xspace}
\newcommand{\appendixtable}[2]{
    \toprule
    \hiderowcolors
    \multicolumn{7}{c}{\textbf{ #1 #2 Score Summary}} \\
    \midrule
        Rank & Name & SQuAD & #1 & Gap & \makecell{New \\Rank} & $\Delta$ Rank \\
    \midrule
    \showrowcolors
    \endfirsthead
    \toprule
    \hiderowcolors
    \multicolumn{7}{c}{\textbf{ #1 #2 Score Summary}} \\
    \midrule
        Rank & Name & SQuAD & #1 & Gap & \makecell{New \\Rank} & $\Delta$ Rank \\
    \midrule
    \showrowcolors
    \endhead
    \multicolumn{7}{c}{Continued on next page} \\
    \endfoot
    \bottomrule
    \endlastfoot}

\gdef\isarxiv{1}
\newcommand{\arxiv}[2]{\ifdefined\isarxiv{#1}\else{#2} \fi }

\date{}

\begin{document}

\title{The Effect of Natural Distribution Shift \\ on Question Answering Models}

\author{
  John Miller \\ UC Berkeley \and
  Karl Krauth \\ UC Berkeley \and
  Benjamin Recht \\ UC Berkeley\and
  Ludwig Schmidt \\ UC Berkeley \and}

\maketitle

\begin{abstract}
We build four new test sets for the Stanford Question Answering Dataset (SQuAD)
and evaluate the ability of question-answering systems to generalize to new
data. Our first test set is from the original Wikipedia domain and measures
the extent to which existing systems overfit the original test set. Despite
several years of heavy test set re-use, we find no evidence of adaptive
overfitting.  The remaining three test sets are constructed from New York
Times articles, Reddit posts, and Amazon product reviews and measure
robustness to natural distribution shifts. Across a broad range of models,
we observe average performance drops of \nytdrop, \redditdrop, and
\amazondrop F1 points, respectively. In contrast, a strong human baseline
matches or exceeds the performance of SQuAD models on the original domain
and exhibits little to no drop in new domains. Taken together, our results
confirm the surprising resilience of the holdout method and emphasize the
need to move towards evaluation metrics that incorporate robustness to natural
distribution shifts.
\end{abstract}

\etocdepthtag.toc{mtsection}

\section{Introduction}
\label{sec:intro}
Since its release in 2016, the Stanford Question Answering Dataset
(\squad)~\citep{rajpurkar2016squad} has generated intense interest from the
natural language processing community. At first glance, this intense interest
has lead to impressive results. The best performing models in
2020~\citep{devlin2018bert, yang2019xlnet} have F1 scores more than 40
points higher than the baseline presented by~\citet{rajpurkar2016squad}. At the same
time, it remains unclear to what extent progress on these benchmark numbers is a
reliable indicator of progress more broadly.

The goal of building question answering systems is not merely to obtain high
scores on the \squad leaderboard, but rather to \emph{generalize} to new
examples beyond the \squad test set. However, the competition format of \squad puts pressure
on the validity of leaderboard scores. It is well-known that repeatedly
evaluating models on a held-out test set can give overly optimistic estimates of
model performance, a phenomenon known as \emph{adaptive
overfitting}~\cite{dwork2015preserving}. Moreover, the standard \squad
evaluation only measures model performance on new examples \emph{from
a single distribution}, i.e., paragraphs derived from Wikipedia articles. 
Nevertheless, we often use models in settings different from the one in which they were trained.
While \citet{jia2017adversarial} demonstrated that \squad models are not robust to
\emph{adversarial} distribution shifts, one might still hope that the models are
more robust to \emph{natural} distribution shifts, for instance changing from
Wikipedia to newspaper articles.

This state of affairs raises two important questions:\\[-.5cm]
\begin{center}
    \emph{Are \squad models overfit to the \squad test set?}\\[.1cm]
    \emph{Are \squad models robust to natural distribution shifts?}
\end{center} 

\begin{figure*}[ht!]
\centering
    \arxiv{
        \begin{subfigure}{\textwidth}
            \centering
            \includegraphics[width=\linewidth]{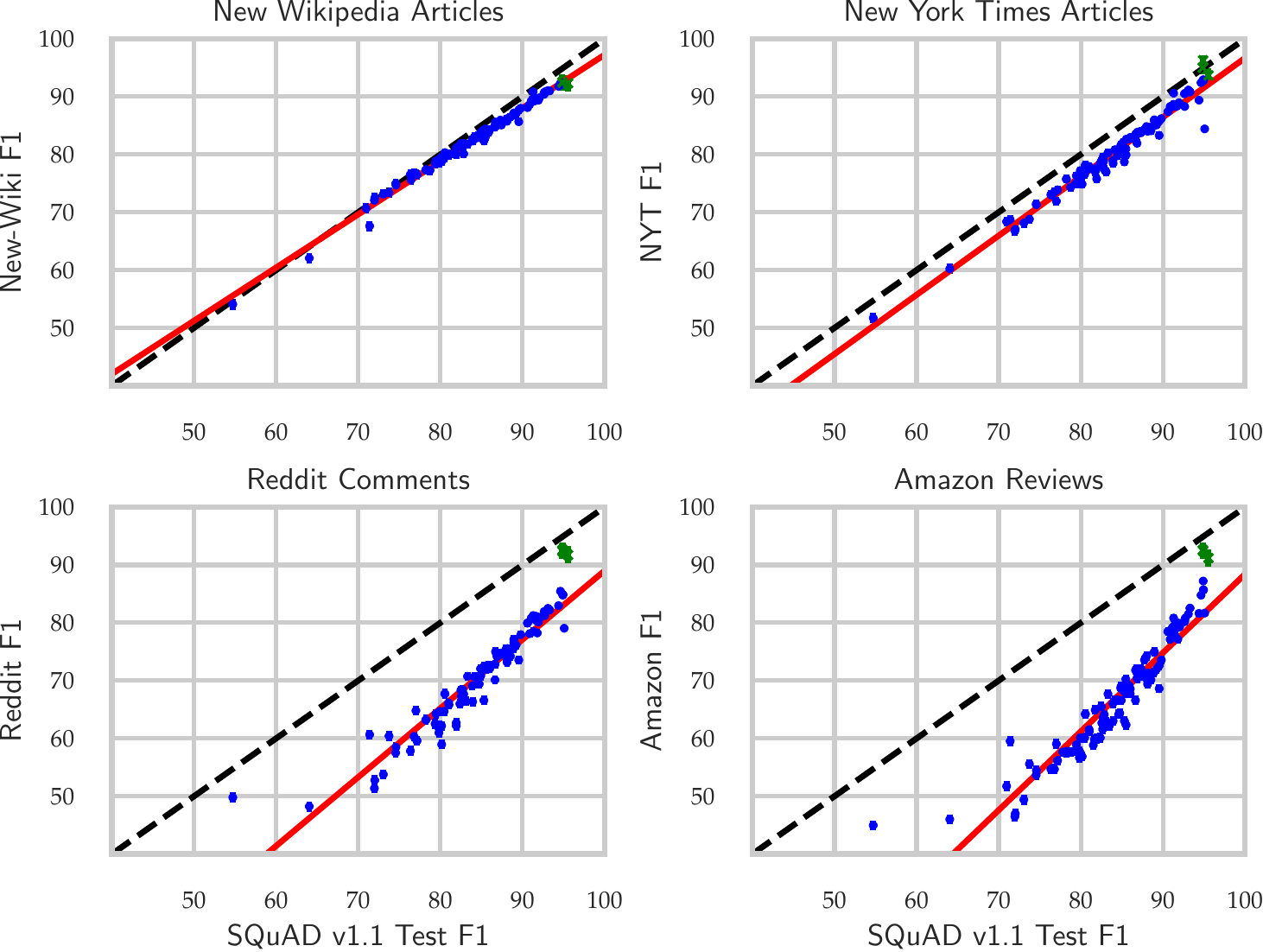}
        \end{subfigure} \\
        \vspace{0.1cm}
        \begin{subfigure}{\textwidth}
            \centering
            \includegraphics{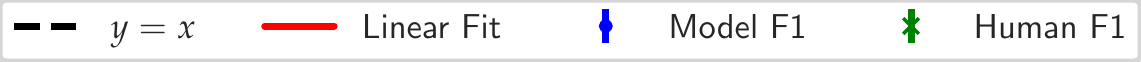}
        \end{subfigure} \\
    }{
        \begin{subfigure}{\textwidth}
            \centering
            \includegraphics[width=\linewidth]{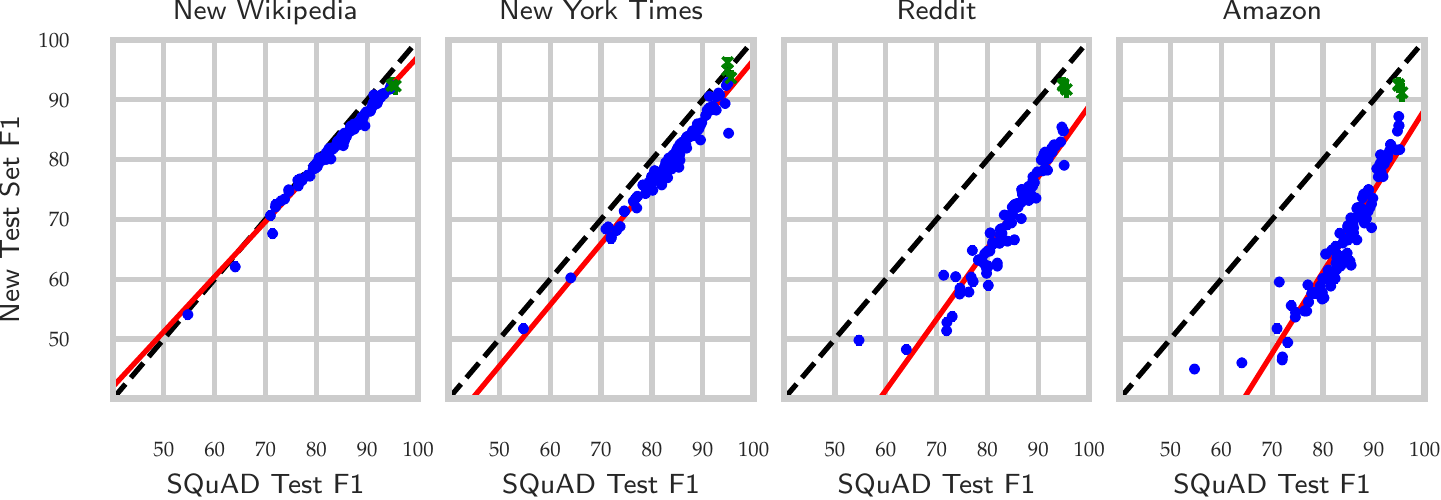}
        \end{subfigure} \\
        \begin{subfigure}{\textwidth}
            \centering
            \includegraphics[width=0.6\linewidth]{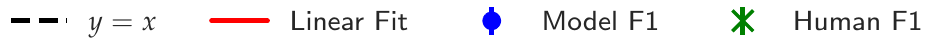}
        \end{subfigure} \\
    }
    \caption{
        Model and human F1 scores on the original \squad v1.1 test set compared
        to our new test sets. Each point corresponds to a model evaluation,
        shown with 95\% Student's t-confidence intervals (mostly covered by the point markers). The plots reveal three
        main phenomena: (i) There is no evidence of adaptive overfitting on
        \squad, (ii) all of the models suffer F1 drops on the new datasets,
        with the magnitude of the drop strongly depending on the corpus, and (iii)
        humans are substantially more robust to natural distribution shifts than
        the models. The slopes of the linear fits are \wikislope, \nytslope,
        \redditslope, and \amazonslope, respectively, and the $R^2$ statistics
        for the linear fits are 0.99, 0.97, 0.9, and 0.89, respectively.  This
        means that every point of F1 improvement on the original dataset
        translates into roughly 1 point of improvement on our new datasets.
    }
\label{fig:main_figure}
\end{figure*}

In this work, we address both questions by replicating the \squad
dataset creation process and generating four new \squad test sets on both the
original Wikipedia domain, as well as three new domains: New York Times
articles, Reddit posts, and Amazon product reviews.

We first show that there is no evidence of adaptive overfitting on \squad.
Across a large collection of \squad models, there is little to no difference
between the F1 scores from the original \squad test set and our replication.
This even holds when comparing scores from the \squad \emph{development} set
(which was publicly released with answers) to our new test set.  The lack of
adaptive overfitting is consistent with recent replication studies in the
context of image classification~\cite{recht2019imagenet,yadav2019cold}. These
studies leave open the possibility that this phenomenon is specific to the data
or models typical in computer vision research. Our result
demonstrates this same phenomenon also holds for natural language processing.

Beyond adaptive overfitting, we also demonstrate that \squad models exhibit
robustness to some of our natural distribution shifts, though they still
suffer substantial performance degradation on others. On the New York Times dataset, 
models in our testbed on average drop \nytdrop F1 points. On the Reddit and Amazon
datasets, the drop is on average \redditdrop and \amazondrop F1 points, respectively. All of
our datasets were collected using the same data generation pipeline, so this
degradation can be attributed purely to changes in the source text rather than
differences in the annotation procedures across datasets.

We complement each of these experiments with a strong human baseline comprised
of the authors of this paper. On the original \squad data, our human accuracy
numbers are on par with the best \squad models~\citep{yang2019xlnet} and
significantly better than the Mechanical Turk baseline reported
by~\citet{rajpurkar2016squad}.  On our new test sets, average human F1 scores
decrease by 0.1 F1 on New York Times, 2.9 on Reddit,
and 3.0 on Amazon.
All of the resulting F1 scores are substantially higher than the best \squad models on the respective test sets.

Figure~\ref{fig:main_figure} summarizes the main results of our experiments.
Humans show consistent behavior on all four test sets, while models are substantially less robust against two of the distribution shifts. 
Although there has been steady progress on the \squad leaderboard, there has been markedly less
progress in this robustness dimension.

To enable future research, all of our new
tests sets are freely available online.\footnote{https://modestyachts.github.io/squadshifts-website/}

\section{Background}
\label{sec:background}
In this section, we briefly introduce the \squad dataset and present a
formal model for reasoning about performance drops between our test sets.

\subsection{Stanford Question Answering Dataset}
\begin{figure}[ht]
    \centering
    \arxiv{
        \includegraphics[height=0.5\textwidth]{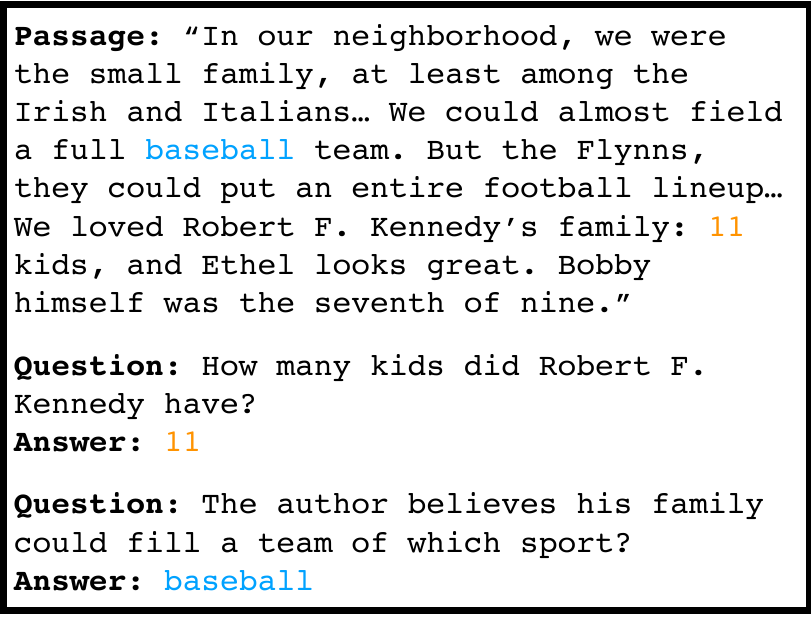}
    }{
        \includegraphics[width=\linewidth]{figures/squad_sample}
    }
    \caption{Question and answer pairs from a sample passage in our New York
    Times \squad test set. Answers are text spans from the passage that answer
    the question.}
\label{fig:squad_sample}
\end{figure}

\squad is an extractive question answering dataset introduced
by~\citet{rajpurkar2016squad}.  An example in \squad consists of a passage of
text, a question, and one or more spans of text within the passage that answer
the question. An example is given in Figure~\ref{fig:squad_sample}.

Model performance is evaluated using one of two metrics: exact match (EM) or F1.
Exact match measures the percentage of predictions that exactly match at least
one of the ground truth answers. F1 measures the maximum overlap between the
tokens in the predicted span and any of the ground truth answers, treating both
the prediction and each answer as a bag of words. Both metrics are described
formally in Appendix~\ref{app:metrics}.

After releasing the original \squad v1.1 dataset, \citet{rajpurkar2018know}
introduced a new variant of the dataset, \squad 2.0, that includes
unanswerable questions.  Since \squad v1.1 has been public for longer and
potentially subjected to more adaptivity, we focus on \squad v1.1 and
refer to it as the \squad dataset throughout our paper. The \squad v1.1 test set is not
publically available. Therefore, while we use public test set evaluation
numbers, we otherwise use the public \squad v1.1 development set for analysis.

\subsection{A Model for Generalization}
Although progress on \squad is measured through performance on a held-out test
set, the implicit goal is not to achieve high F1 scores on the test set, but
rather to \emph{generalize} to unseen examples. Our experiments test the extent
to which this assumption holds---if models with high leaderboard scores on
the test set continue to perform well on new examples, whether from the same or
different distributions.

To be more formal, suppose the original test set $S$ is sampled from
some underlying distribution $\dist$, and consider a model $f$ submitted to
the \squad leaderboard. Let $L_S(f)$ denote the empirical loss of model $f$ on
the sample $S$, and let $L_\dist(f)$ denote the corresponding population loss.
In our experiment, we gather a new dataset of examples $S'$ from a distribution
$\dist'$, potentially different from $\dist$.  We wish for the loss on the new
sample, $L_{S'}(f)$ to be close to the original, $L_S(f)$. Omitting $f$, we can
decompose this gap into three terms~\citep{recht2019imagenet}.
\begin{align*}
    L_S  - L_{S'} 
    = \underbrace{(L_S - L_\dist)}_{\text{Adaptivity gap}}
    + \underbrace{(L_\dist - L_{\dist'})}_{\text{Distribution gap}}
    + \underbrace{(L_{\dist'} - L_{S'})}_{\text{Generalization gap}}
\end{align*}

The \emph{adaptivity gap} $L_S - L_\dist$ measures how much adapting the model
to the held-out test set $S$ biases the estimate of the population loss. Since recent
models are in part chosen on the basis of past test set information, the model
$f$ is not independent of $S$. Hence $L_S(f)$ can underestimate $L_\dist(f)$, a
phenomenon called \emph{adaptive overfitting.}  The \emph{distribution gap}
measures how much changing the distribution from $\dist$ to $\dist'$ affects the
model's performance. Finally, the \emph{generalization gap} $L_{S'} - L_{\dist'}$
captures the difference between the sample and the population losses due to
random sampling of $S'$. Since $S'$ is sampled independently of the model $f$, this gap is
typically small and well-controlled by standard concentration results. 
For example, on the new Wikipedia test set, the average size of Student's t-confidence intervals for models in our testbed is $\pm 0.6$ F1.

In the sequel, we empirically measure both the adaptivity gap and the
distribution gap for a wide range of \squad models by collecting new test
sets from a variety of distributions $\dist'$. We first review related work that
motivates our choice of \squad and natural distribution shifts.

\section{Related Work}
\label{sec:related_work}

\paragraph{Adaptive data analysis.}
Although repeated test-set reuse puts pressure on the statistical guarantees of
the holdout method~\cite{dwork2015preserving}, a series of replication studies established there is no
adaptive overfitting on popular classification benchmarks like
MNIST~\citep{yadav2019cold}, CIFAR-10~\citep{recht2019imagenet}, and
ImageNet~\citep{recht2019imagenet}.
Furthermore, \citet{roelofs2019meta} also found little to no evidence of
adaptive overfitting in a host of classification competitions on the Kaggle
platform. These investigations either concern
image classification or smaller competitions that have not been subject to intense,
multi-year community scrutiny. Our work establishes similar results
for natural language processing on a heavily studied benchmark. 

A number of works have proffered
explanations for why adaptive overfitting does not occur in the standard machine
learning workflow~\citep{blum2015ladder, mania2019model, feldman2019advantages, zrnic2019natural}.
Complementary to these results, our work provides a new data point with which to
validate and deepen our conceptual understanding of overfitting.

\paragraph{Datasets for question answering.}
Beyond \squad, a number of works have proposed datasets for question
answering~\citep{richardson2013mctest,berant2014modeling,joshi2017triviaqa,trischler2017newsqa,dunn2017searchqa,yang2018hotpotqa,
kwiatkowski2019natural}.  We focus our analysis on \squad for two reasons.
First, \squad has been the focus of intense research for almost four years, and
the competitive nature of the leaderboard format makes it an excellent example
to study adaptive overfitting in natural language processing. Second, \squad
requires all submissions to be uploaded to
CodaLab\footnote{\url{https://worksheets.codalab.org/}}, which ensures reproducibility
and makes it possible to evaluate every submission on our new datasets using the
same configuration and environment as the original evaluation.

\paragraph{Generalization in question answering.}
Given the plethora of question-answering datasets, \citet{yogatama2019learning},
\citet{talmor2019multiqa}, and \citet{sen2020models} evaluate the extent to
which models trained on \squad generalize to other question-answering datasets.
\citet{hendrycks2020pretrained} evaluates generalization under distribution
shift for question answering, among other tasks, by carefully splitting subsets
of the ReCoRD~\cite{zhang2018record} dataset. In a similar vein,
\citet{fisch2019mrqa} conduct a shared task competition that evaluates how well
models trained on a collection of six datasets generalize to unseen datasets at
test time.  In these cases, the datasets encountered at test time vary across a
number of dimensions: the question collection procedure, the origin of the input
text, the question answering interface, the crowd worker population, etc. These
differences are \emph{confounding factors} that make it difficult to interpret
performance differences across datasets. For example, human performance differs
by 10 F1 points between \squad v1.1 and NewsQA~\citep{trischler2017newsqa}. In
contrast, our datasets focus on a single factor of variation---the input text
corpus. In this controlled setting, we observe non-trivial F1 drops across a
large collection of models, while human F1 scores are essentially constant.

From a different perspective, \citet{jia2017adversarial} and
\citet{ribeiro2018semantically} consider robustness to \emph{adversarial}
dataset corruptions. \citet{kaushik2019learning} and
\citet{gardner2020evaluating} evaluate model performance when individual examples are
perturbed in small, but semantically meaningful ways. While we instead focus on
\emph{naturally occurring} distribution shifts, we also evaluate our model
testbed on adversarial distribution shifts for comparison in
Appendix~\ref{app:adversarial}.

\section{Collecting New Test Sets}
\label{sec:data-collection}

In this section, we describe our data collection methodology. Data collection
primarily proceeds in two stages: curating passages from a text corpus and
crowdsourcing question-answer pairs over the passages. In both of these stages,
we take great care to replicate the original \squad data generation process.
Where possible, we obtained and used the original \squad generation code
kindly provided by~\citet{rajpurkar2016squad}. We ran our dataset creation pipeline on four
different corpora: Wikipedia articles, New York Times articles, Reddit posts,
and Amazon product reviews.

\subsection{Passage Curation}
The first step in the dataset generation process is selecting the
articles from which the passages or contexts are drawn.

\paragraph{Wikipedia.}
We sampled 48 articles uniformly at random from the same list of 10,000
Wikipedia articles as~\citet{rajpurkar2016squad}, ensuring that there is no overlap
between our articles and those in the \squad v1.1 training or development
sets. To minimize distribution shift due to temporal language variation, we
extracted the text of the Wikipedia articles from around the publication date of
the \squad v1.0 dataset (June 16, 2016). For each article, we extracted
individual paragraphs and stripped out images, figures, and tables using the
same data processing code as~\citet{rajpurkar2016squad}. Then, we subsampled the
resulting paragraphs to match the passage length statistics of the original
\squad dataset.\footnote{The minimum 500 character per paragraph rule mentioned
in \citet{rajpurkar2016squad} was adopted midway through their data collection, and
hence the original dataset also includes shorter paragraphs
\citep{rajpurkar2019communication}.} See Appendix~\ref{app:length_statistics}
for a detailed comparison of the paragraph distribution of the
original SQuAD dev set and our new SQuAD test set.

\paragraph{New York Times.}
We sampled New York Times articles from the set of all articles published in
2015 using the NYTimes Archive API. We scraped each article with the Wayback
Machine\footnote{\url{https://archive.org/web/}}, 
using the same snapshot timestamp as our Wikipedia dataset, and
removed foreign language articles.  Since the average paragraph length for NYT
articles is significantly shorter than the average paragraph length for
Wikipedia articles, we merged each NYT paragraph with its subsequent
paragraph with some probability.
Then we subsampled the merged paragraphs to match the
passage length statistics of the original \squad v1.1 dataset.

\paragraph{Reddit Posts.}
We sampled Reddit posts from the set of all posts across all subreddits during the month of January 2016 in the Pushshift Reddit
Corpus~\citep{baumgartner2020pushshift}.
Then we restricted the set of posts to those marked as ``safe for work'' and manually removed inappropriate posts from the remaining ones.
We concatenated each post's title
with its body, removed Markdown, and replaced all links with a single token, {\tt
LINKREMOVED}. We then subsampled the posts to match the passage length
statistics of the original \squad v1.1 dataset.

\paragraph{Amazon Product Reviews.}
We sampled Amazon product reviews belonging to the ``Home and Kitchen'' category from
the dataset released by~\citet{mcauley2015image}. As in the previous datasets,
we then subsampled the reviews to match the passage length statistics of 
\squad v1.1.

\begin{table}
    \centering
    \caption{Dataset statistics of our four new test sets compared to the
    original SQuAD 1.1 development and test sets.}
    \rowcolors{2}{white}{gray!15}
\begin{tabular}{lcc}
\toprule 
Dataset & Total Articles & Total Examples \\
\midrule
    SQuAD v1.1 Dev & 48 & 10,570 \\
    SQuAD v1.1 Test & 46 & 9,533 \\
    New Wikipedia & 48 & 7,938 \\
    New York Times & 797 & 10,065 \\
    Reddit & 1969 & 9,803 \\
    Amazon & 1909 & 9,885 \\
\bottomrule
\end{tabular}

    \label{table:dataset_summary}
\end{table}

\subsection{Crowdsourcing Question-Answer Pairs}
We employed crowdworkers on Amazon Mechanical Turk (MTurk) to ask and answer
questions on the passages in each dataset. We followed a nearly identical
protocol to the original \squad dataset creation process. We used the same MTurk
user interface, task instructions, MTurk worker qualifications, time per task,
and hourly rate (adjusted for inflation) as~\citet{rajpurkar2016squad}.
For full details and examples of the user interface, refer to
Appendix~\ref{app:mturk_details}.

For each paragraph, one crowdworker first asked and answered up
to five questions on the content of the paragraph. Then we obtained at least two
additional answers for each question using separate crowdworkers.
There are two points of discrepancy between our crowdsourcing protocol and the
one used to create the original \squad dataset. First, we interfaced directly
with MTurk rather than via the Daemo platform because the Daemo platform has
been discontinued. Second, in our MTurk tasks, workers asked and answered questions
for at most five paragraphs rather than for the entire article because MTurk
workers preferred smaller units of work. Although each difference is a potential source
of distribution shift, in Section~\ref{sec:results} we show that the effect of
these changes is negligible---models achieve roughly the same scores on both
the original and new Wikipedia datasets. On average, the difference in F1 scores
is 1.5 F1, and 95\% of models in our testbed are within 2.7 F1.

After gathering question and answer pairs for each paragraph, we apply the same
post-processing and data cleaning as \squad v1.1. We adjusted answer whitespace
for consistency, filtered malformed answers, and removed all documents that had
less than an average of two questions per paragraph after filtering. In
Appendix~\ref{app:data_cleaning}, we show that further manual filtering of
incorrect, ungrammatical, or otherwise malformed questions and answers has
negligible impact on our results. Table~\ref{table:dataset_summary} summarizes
the overall statistics of our datasets.

\subsection{Human Evaluation}
Although both \squad and our new test sets have answers from MTurk workers,
it is not clear whether these answers represent a compelling human baseline. At
minimum, workers are not familiar with the typical style of answers in \squad
(e.g., how much detail to include), and they receive no feedback on their
performance. To obtain a stronger human baseline, the graduate student and
postdoc authors of this paper also answered approximately 1,000 questions on
each of the four new test sets and the original \squad development set,
following the same procedure and using the same UI as the MTurk workers. To take
feedback into account, each participant first labelled 500 practice examples
from the training set and compared their answers with the ground truth.

\section{Main Results}
\label{sec:results}
We use the four new datasets generated in the previous part to test for adaptive
overfitting on \squad and probe the robustness of \squad models to natural
distribution shifts. 

We evaluated a broad set of over 100 models submitted to the \squad
leaderboard, including state-of-the-art models like XLNet~\citep{yang2019xlnet}
and BERT~\citep{devlin2018bert}, as well as older, but popular models like
BiDAF~\citep{seo2016bidirectional}. All of the models were submitted to the
CodaLab platform, and we evaluate every model using the exact same configuration
(model weights, hyperparameters, command-line arguments, execution environment)
as the original submission.  
\arxiv{
Tables~\ref{table:new_wiki_f1_table} and~\ref{table:amazon_f1_table} contain
a brief summary of the results for key models. 
}{
Table~\ref{table:main_f1_table} contains a brief summary of the results for key models.}
Detailed results table and citations for the models, where available, are
given in Appendix~\ref{app:full_results}.

\arxiv{}
{
    \begin{table*}[t!]
    \centering
    \caption{Model F1 scores on the original SQuAD test set and our new test sets
        from Wikipedia, New York Times, Reddit, and Amazon data. Rank refers to the
        original public leaderboard rank. The confidence intervals are 95\%
        Student's t-intervals. A complete table with data for the entire model
        testbed, references, and analogous data for EM scores is in
        Appendix~\ref{app:full_results}.}
        \rowcolors{2}{gray!15}{white}
\begin{tabular}{lccccc}
    \toprule
    \multicolumn{6}{c}{\textbf{F1 Score Summary}} \\
    \midrule
        Name & SQuADv1.1 & New-Wiki & NYT & Reddit  & Amazon  \\
    \midrule
	Human (average) & - & 95.1 & 92.4& 95.0 & 92.2 & 92.1\\
	XLNET-123 & 2 & 94.9 & 92.2  \textcolor{gray!70}{[91.7, 92.7]} & 92.8  \textcolor{gray!70}{[92.3, 93.2]} & 84.9  \textcolor{gray!70}{[84.2, 85.5]} & 85.7  \textcolor{gray!70}{[85.1, 86.3]} \\
	Tuned BERT-1seq Large & 5 & 93.3 & 91.0  \textcolor{gray!70}{[90.5, 91.5]} & 90.8  \textcolor{gray!70}{[90.3, 91.3]} & 82.2  \textcolor{gray!70}{[81.5, 82.9]} & 82.5  \textcolor{gray!70}{[81.9, 83.2]} \\
	BERT-Large Baseline & 7 & 92.7 & 90.8  \textcolor{gray!70}{[90.3, 91.3]} & 90.6  \textcolor{gray!70}{[90.1, 91.1]} & 81.2  \textcolor{gray!70}{[80.6, 81.9]} & 80.8  \textcolor{gray!70}{[80.2, 81.5]} \\
	BiDAF+SelfAttention+ELMo & 25 & 85.9 & 83.8  \textcolor{gray!70}{[83.1, 84.5]} & 82.7  \textcolor{gray!70}{[82.0, 83.4]} & 72.6  \textcolor{gray!70}{[71.8, 73.4]} & 69.2  \textcolor{gray!70}{[68.3, 70.0]} \\
	Jenga & 38 & 82.8 & 80.1  \textcolor{gray!70}{[79.3, 80.9]} & 77.4  \textcolor{gray!70}{[76.7, 78.1]} & 67.7  \textcolor{gray!70}{[66.8, 68.5]} & 64.1  \textcolor{gray!70}{[63.3, 65.0]} \\
	AllenNLP BiDAF & 53 & 77.2 & 76.5  \textcolor{gray!70}{[75.7, 77.3]} & 73.8  \textcolor{gray!70}{[73.1, 74.6]} & 59.6  \textcolor{gray!70}{[58.7, 60.4]} & 56.2  \textcolor{gray!70}{[55.3, 57.0]} \\
	\bottomrule
\end{tabular}

        \label{table:main_f1_table}
    \end{table*}
}

\subsection{Adaptive Overfitting}
\label{sec:results:adaptive}

\arxiv{
    \begin{table*}
        \centering
        \caption{
            Comparison of model F1 scores on the original SQuAD test set and
            our new Wikipedia test set. Rank refers to the relative ordering of
            the models in our testbed using the original \squad v1.1 F1
            scores, new rank refers to the ordering using the new Wikipedia test
            set scores, and $\Delta$ rank is the relative difference in ranking
            from the original test set to the new test set.  The confidence
            intervals are 95\% Student's t-intervals. No confidence intervals are
            provided for the \squad v1.1 dataset since the dataset is not public
            and only the average scores are available. A complete table with data
            for the entire model testbed, references, and analogous data for EM
            scores is in Appendix~\ref{app:full_results}.
        }
        \rowcolors{2}{gray!15}{white}
    \begin{tabular}{llccccc}
        \toprule
        \multicolumn{7}{c}{\textbf{ New-Wiki F1 Score Summary}} \\
        \midrule
            Rank & Name & SQuAD & New-Wiki& Gap & \makecell{New \\Rank} & $\Delta$ Rank \\
        \midrule
	- & Human average (this study) & 95.1 & 92.4& 2.7 & - & -\\
	1 & XLNet & 95.1 & 92.3  \textcolor{gray!90}{[91.9, 92.8]} & 2.7 & 1 & 0 \\
	2 & XLNET-123 & 94.9 & 92.2  \textcolor{gray!90}{[91.7, 92.7]} & 2.7 & 4 & -2 \\
	6 & Tuned BERT-1seq Large & 93.3 & 91.0  \textcolor{gray!90}{[90.5, 91.5]} & 2.3 & 7 & -1 \\
	8 & BERT-Large Baseline & 92.7 & 90.8  \textcolor{gray!90}{[90.3, 91.3]} & 1.9 & 9 & -1 \\
	42 & BiDAF+SelfAttention+ELMo & 85.9 & 83.8  \textcolor{gray!90}{[83.1, 84.5]} & 2.1 & 45 & -3 \\
	62 & Jenga & 82.8 & 80.1  \textcolor{gray!90}{[79.3, 80.9]} & 2.7 & 71 & -9 \\
	85 & AllenNLP BiDAF & 77.2 & 76.5  \textcolor{gray!90}{[75.7, 77.3]} & 0.7 & 88 & -3 \\
	\bottomrule
\end{tabular}

        \label{table:new_wiki_f1_table}
    \end{table*}
}{}

The \squad models in our testbed come from a long sequence of papers that
incrementally improve F1 and EM scores over a period of several years.
Consequently, if there is adaptive overfitting, we should expect the later
models to have larger drops in F1 scores  because they are the result of more
interaction with the test set. In this case, the higher F1 scores are partially
the result of a larger adaptivity gap, and we would expect that, as the observed
scores $L_S$ continue to rise, the population scores $L_D$ would begin to
plateau.

To check for adaptive overfitting on the existing test set, we plot the \squad v1.1 test F1 scores
against F1 scores on our new Wikipedia test set.  Figure~\ref{fig:main_figure} in
Section~\ref{sec:intro} provides strong evidence against the adaptive
overfitting hypothesis. Across the entire model collection, the F1 scores on the
new test set closely replicate the original F1 scores. The observed linear fit
is in contrast to the concave curve one would expect from adaptive overfitting.
We use 95\% Student's t-confidence intervals, which make a large-sample Gaussian
assumption, to capture the error in the new F1 scores due to random
variation. No such confidence intervals are available for the original test set
scores since the test set is not publicly available.  A similar plot for EM
scores is provided in Appendix~\ref{app:scatterplots}.

\begin{figure}[t!]
\centering
    \arxiv{
        \includegraphics[width=0.7\textwidth]{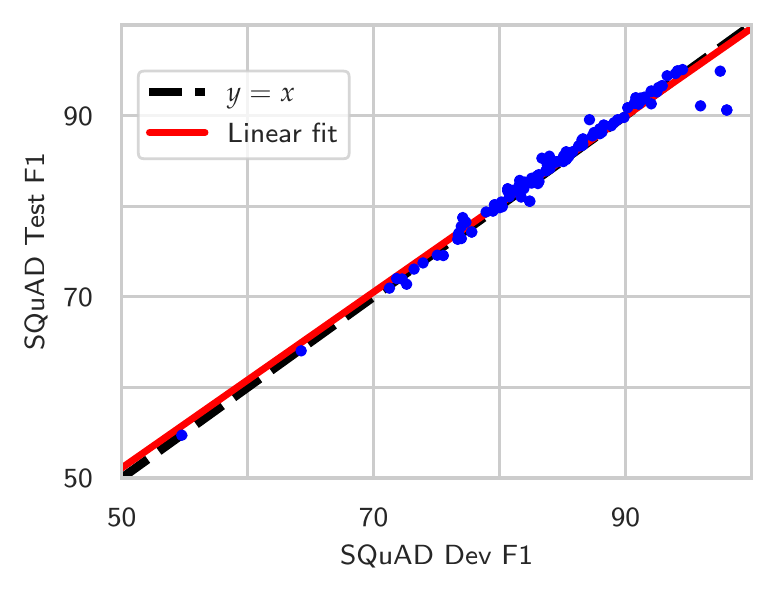}
    }{
        \includegraphics[width=\linewidth]{figures/dev_vs_test}
    }
\caption{Comparison of F1 scores between the \squad v1.1 dev set and the
    \squad v1.1 test set. Despite heavy use of the dev set during model development,
    the dev set and test set scores closely match, with the exception of three
    models that were explicitly trained on the dev set, {\tt Common-sense
    Governed BERT-123 (April 21)}, {\tt Common-sense Governed BERT-123 (May 9)},
    and {\tt XLNet-123++}.~\citep{qiu2020communication}.
    The slope of the linear fit is 0.97.}
\label{fig:dev_vs_test}
\end{figure}

Not only is there little evidence for adaptive overfitting on the test set,
there is also little evidence of adaptive overfitting on the \squad development set. In
Figure~\ref{fig:dev_vs_test}, we plot F1 scores on the \squad v1.1 development
set against F1 scores on the \squad v1.1 test set. With the exception of
three models, the F1 scores on the dev set closely match the scores on the test
set, despite the fact that the development set is aggressively used during model
selection. Moreover, the models that do not lie on the linear trend line---{\tt
Common-sense Governed BERT-123 (April 21)}, {\tt Common-sense Governed BERT-123
(May 9)}, and {\tt XLNet-123++}---are directly trained on the development
set~\citep{qiu2020communication}.

\begin{figure}[t!]
    \centering
    \begin{subfigure}{\textwidth}
        \centering
        \includegraphics[width=\linewidth]{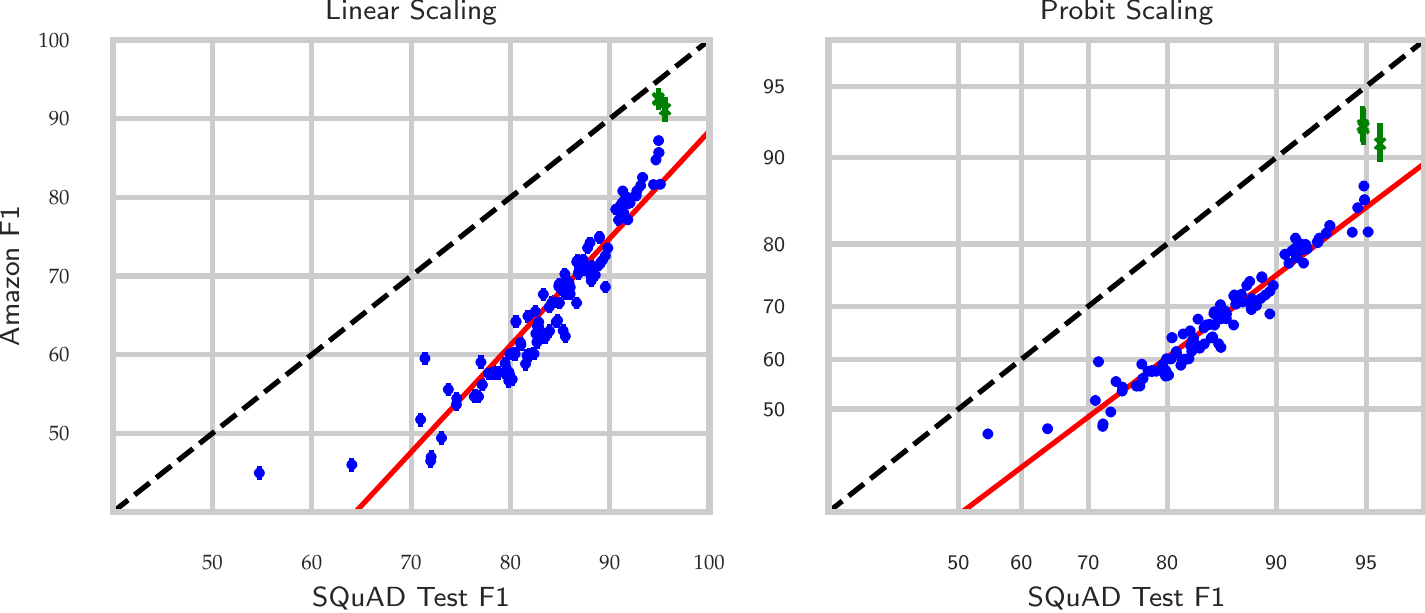}
    \end{subfigure} \\
    \vspace{0.05cm}
    \begin{subfigure}{\textwidth}
        \centering
        \includegraphics[width=0.65\linewidth]{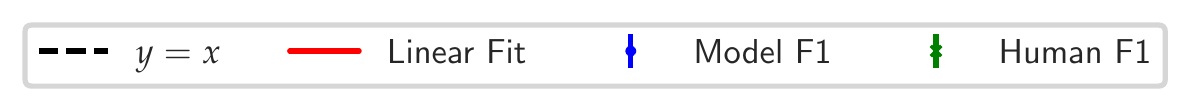}
    \end{subfigure} \\
    \caption{Comparison of model and human F1 scores on the original \squad v1.1
    test set and our new Amazon test set. Each datapoint corresponds to one
    model in the testbed and is shown with 95\% Student's t-confidence
    intervals. The left plot shows the model F1 scores under a linear axis scaling,
    whereas the right plot uses an \emph{probit scale} on both axes. In other words,
    model F1 score $x$ appears at $\Phi^{-1}(x)$, where $\Phi^{-1}$ is the
    inverse Gaussian CDF. Visual inspection shows the linear fit is better
    in the probit domain. Quantitatively, the $R^2$ statistic is $0.89$ in the
    linear domain, compared to $0.94$ in the probit domain. 
    See Appendix~\ref{app:probit_fits} for similar comparisons for all datasets.}
    \label{fig:comparison_probit}
\end{figure}


\subsection{Robustness to Natural Distribution Shifts}
\label{sec:results_shifts}
\arxiv{
    \begin{table*}[t!]
        \centering
        \caption{
            Comparison of model F1 scores on the original SQuAD test set and
            our new Amazon test set. Rank refers to the relative ordering of
            the models in our testbed using the original \squad v1.1 F1
            scores, new rank refers to the ordering using the Amazon test
            set scores, and $\Delta$ rank is the relative difference in ranking
            from the original test set to the new test set.  The confidence
            intervals are 95\% Student's t-intervals.  No confidence intervals are
            provided for \squad v1.1 since the dataset is not public
            and only the average scores are available for each model.  A
            complete table with data for the entire model testbed, the New York
            Times and Reddit datasets, and EM scores scores is in
            Appendix~\ref{app:full_results}.
        }
        \rowcolors{2}{gray!15}{white}
    \begin{tabular}{llccccc}
        \toprule
        \multicolumn{7}{c}{\textbf{ Amazon F1 Score Summary}} \\
        \midrule
            Rank & Name & SQuAD & Amazon& Gap & \makecell{New \\Rank} & $\Delta$ Rank \\
        \midrule
	- & Human average (this study) & 95.1 & 92.1& 3.0 & - & -\\
	1 & XLNet & 95.1 & 81.7  \textcolor{gray!90}{[81.1, 82.2]} & 13.4 & 5 & -4 \\
	2 & XLNET-123 & 94.9 & 85.7  \textcolor{gray!90}{[85.1, 86.3]} & 9.2 & 2 & 0 \\
	6 & Tuned BERT-1seq Large & 93.3 & 82.5  \textcolor{gray!90}{[81.9, 83.2]} & 10.8 & 4 & 2 \\
	8 & BERT-Large Baseline & 92.7 & 80.8  \textcolor{gray!90}{[80.2, 81.5]} & 11.9 & 8 & 0 \\
	45 & BiDAF+SelfAttention+ELMo & 85.9 & 69.2  \textcolor{gray!90}{[68.3, 70.0]} & 16.7 & 43 & 2 \\
	67 & Jenga & 82.8 & 64.1  \textcolor{gray!90}{[63.3, 65.0]} & 18.7 & 65 & 2 \\
	93 & AllenNLP BiDAF & 77.2 & 56.2  \textcolor{gray!90}{[55.3, 57.0]} & 21.0 & 95 & -2 \\
	\bottomrule
\end{tabular}

        \label{table:amazon_f1_table}
    \end{table*}
}{}

Given the correspondence between the old and new Wikipedia test set F1 scores,
the adaptivity gap and the distribution gap are small or non-existent.
Consequently, the distribution shift stemming from our data
generation pipeline affects the models only minimally. This allows us to probe the sensitivity of the \squad
models to a set of controlled distribution shifts, namely the choice of text
corpus. Since all of the datasets are constructed with the same preprocessing
pipeline, crowd-worker population, and post-processing, the datasets are
free of confounding factors that would otherwise arise when comparing model
performance across different datasets.

Figure~\ref{fig:main_figure} in Section~\ref{sec:intro} shows F1 scores
on the \squad v1.1 test set versus the F1 scores on each of our new
test sets for all the models in our testbed. All models experience an F1 drop on the new test sets,
though the magnitude strongly depends on the specific test set.
On New York Times, for instance, BERT only drops around 2.1 F1
points, whereas it drops around 11.9 F1 points on Amazon and 11.5 F1 points on
Reddit. The top performing XLNet model~\citep{yang2019xlnet} is a clear outlier.
Despite generalizing well to the new Wikipedia dataset, XLNet drops nearly 10 F1
and 40 EM points on New York Times, substantially more than models with
similar performance on \squad v1.1 as well as other XLNet variants, e.g.,
XLNet-123\footnote{This large drop persists even when normalizing Unicode
characers and replacing Unicode punctuation with Ascii approximations.}.

\arxiv{
    Table~\ref{table:amazon_f1_table} summarizes the F1 scores for a select set
    of models. Full results for all models, datasets, and EM scores are given in
    Appendix~\ref{app:full_results}. 
}{}

In general, F1 scores on the original \squad test set are highly predictive of
F1 scores on the new test sets. Interestingly, the relationship is
well-captured by a linear fit even under distribution shifts. Similar
to~\citet{recht2019imagenet}, in Figure~\ref{fig:comparison_probit}, we observe
the linear fits are better under a probit scaling of F1 scores. See 
Appendix~\ref{app:probit_fits} for more details. Moreover, the gap between
perfect robustness ($y=x$) and the observed linear fits varies with the dataset:
\nytdrop F1 points for New York Times, \redditdrop points for Reddit, and
\amazondrop F1 for Amazon. In each case, however, higher performance on \squad v1.1
translates into higher performance on these natural distribution shift
instances. 

Despite the robustness demonstrated by the models, on all of the 
test sets with distribution shift, human performance is substantially higher than model performance and
well above the linear fits shown in Figure~\ref{fig:main_figure} and
Figure~\ref{fig:comparison_probit}. This rules out the possibility that the
shift in F1 scores are entirely by a change in the Bayes error rate.
Moreover, it points towards substantial room for improvement for models on our
new test sets. 

\section{Further Analysis}
In this section, we further explore the properties of our new test sets.  We
first study the extent to which common measures of dataset difficulty can explain
the performance drops on our new test sets.  Then, we evaluate whether training models
with more data or more diverse data improves robustness to our distribution
shifts.

\label{sec:analysis}

\subsection{Are The New Test Sets Harder Than The Original?}
One hypothesis for the performance drops observed in
Section~\ref{sec:results_shifts} is that our new dataset are harder in some
sense. For instance, the diversity of answers may be greater among Reddit
comments than Wikipedia articles. To better understand this question, we compare the
original \squad development set to our four new test sets using the three
difficulty measures introduced in \citet{rajpurkar2016squad}.

\begin{figure}[ht!]
    \begin{center}
    \arxiv{
        \includegraphics[width=0.7\textwidth]{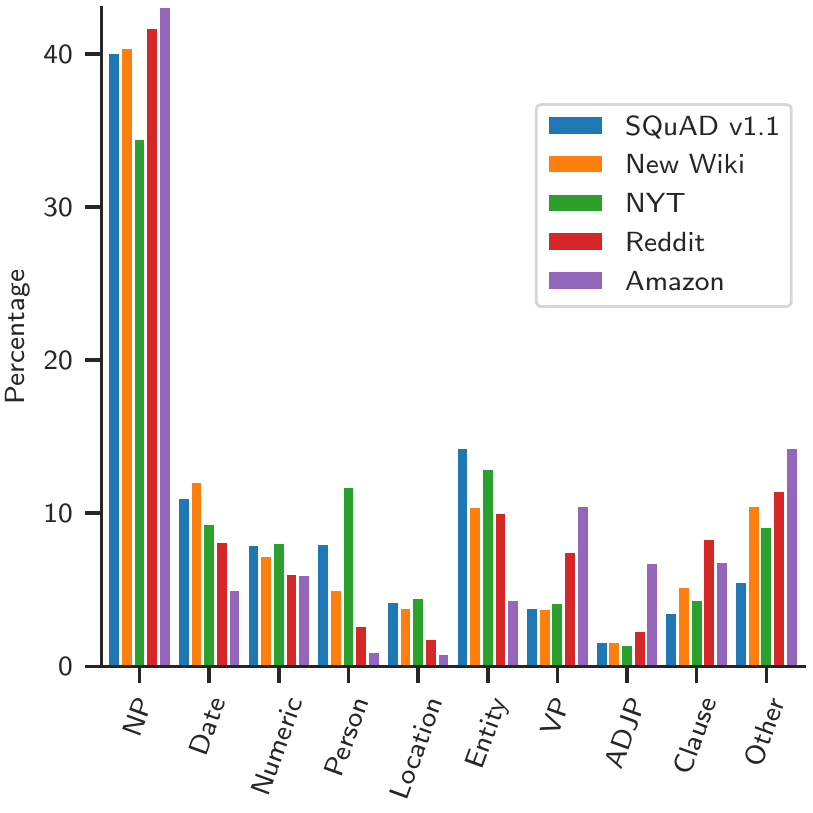}
    }{
        \includegraphics[width=\linewidth]{figures/answer_categorization}
    }
    \end{center}
    \caption{Comparison of answers types in the original and new
    datasets. We automatically partition our answers into the same categories as
    \citet{rajpurkar2016squad}. Although there are differences between the
    datasets, e.g., New York Times has more person answers, the four datasets are
    very similar. Moreover, we show in Appendix~\ref{app:answer_shifts} that 
    differences in answer categorization across datasets do not explain the
    performance drops we observe.}
    \label{fig:answer_categorization}
\end{figure}

\paragraph{Answer diversity.}
Following \citet{rajpurkar2016squad}, we automatically categorize each answer
into numerical and non-numerical answers, named entities, and constituents using
spaCy \citep{spacy2} and the constituency parser from
\citet{kitaev2018selfAttentive}. Histograms of answer types for each data are
shown in Figure~\ref{fig:answer_categorization}. 
Since the original pipeline is not available, our implementation differs
slightly from \citet{rajpurkar2016squad} and we include results on the 
\squad v1.1 development set for comparison. Both the original and our new
Wikipedia test set have very similar answer type histograms. The distribution
shift datasets have slight variations in the answer distributions. For instance,
NYT has more person answers, whereas Amazon has more adjective phrases. However,
changes in the answer type distribution between datasets are not sufficient to
explain the performance differences between the datasets. In
Appendix~\ref{app:answer_shifts}, we consider a simple model that predicts F1
scores on our new test sets by stratifying the dataset by answer type, computing
model F1 scores for each type, and then reweighing these scores by the relative
frequency of each answer type in our new test set. This model explains only a
small fraction of the performance differences across test sets.

\begin{figure}[ht!]
    \center
    \arxiv{
        \includegraphics[width=0.7\textwidth]{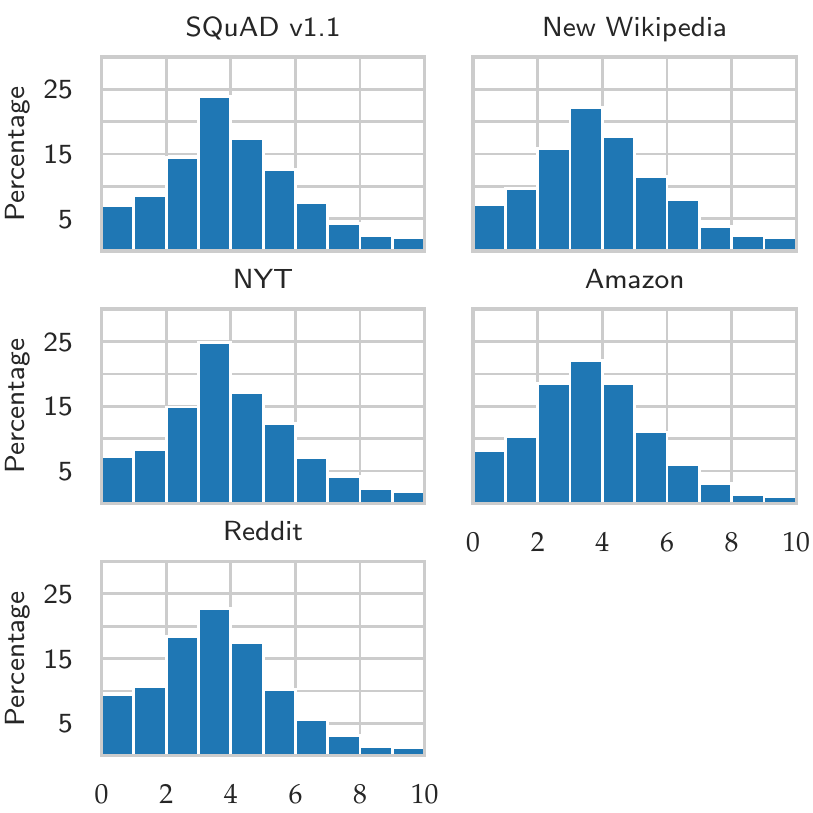}
    }{
        \includegraphics[width=\linewidth]{figures/syntactic_divergence}
    }
    \caption{Histograms of syntactic divergence between question and answer
    sentences for both the original and new datasets. All of the datasets have a
    similar distribution of syntactic divergence, though the Reddit and Amazon
    datasets have more question-answers pairs with small (1-2) syntactic
    divergence.}
\label{fig:syntactic_divergence}
\end{figure}

\paragraph{Syntactic divergence.}
We also stratify our datasets using the automatic syntactic divergence measure
of \citet{rajpurkar2016squad}. Syntactic divergence measures the similarity
between the syntactic dependency tree structure of both the question and answer
sentences and provides another metric of example difficulty. In
Figure~\ref{fig:syntactic_divergence}, we compare the histograms of syntactic
divergence for the \squad v1.1 development set and our new test sets. All of the
datasets have similar histograms, though both the Reddit and Amazon test sets
have slightly more examples with small syntactic divergence. As in the previous
part, in Appendix~\ref{app:syntax_shifts}, we consider a simple model that
predicts F1 scores on the new test sets by stratifying the dataset according to
syntactic divergence and reweighting based on the relative frequence of examples
with a given syntactic divergence measure. As before, this model explains only a
small fraction of the performance differences across test sets.

\paragraph{Reasoning required.}
Finally, we compare our new test sets in terms of the reasoning required to
answer each question-answer pair, using the same non-mutually exclusive
categories as~\citet{rajpurkar2016squad}. For each test set, as well as the
\squad development set, we randomly sampled and manually labeled 192 examples.
The results for each dataset are presented in
Table~\ref{table:reasoning_required}.  Both the Amazon and Reddit dataset have
more examples requiring world knowledge to resolve lexical variation, while the
New York Times dataset has more examples requiring multi-sentence reasoning.
Differences in reasoning required between test sets do not explain the
observed performance drops. In Appendix~\ref{app:reasoning_shifts}, we present a
another model that predicts F1 scores on our new test sets by computing model F1
scores in each reasoning category and then reweighing these scores based on the
relative frequency of each category on new test sets. This model explains
virtually none of the observed changes in F1 scores.

\begin{table*}[ht]
    \centering
    \caption{Manual comparison of the reasoning required to answer each question-answer
    pair on a random sample of 192 examples from each dataset using the
    categories from~\citet{rajpurkar2016squad}. The Reddit and Amazon datasets
    have more examples requiring world knowledge to resolve lexical
    variation, whereas the New York Times and Amazon datasets require more multi-sentence
    reasoning. We show in Appendix~\ref{app:reasoning_shifts} that these
    differences in reasoning required do not explain the performance drops we observe.}
    \rowcolors{2}{white}{gray!15}
\begin{tabular}{lccccc}
\toprule 
	 Reasoning Type & SQuAD v1.1& New Wiki & NYT & Reddit & Amazon \\
\midrule
	Lexical Variation (Synonomy) & 39.1 & 39.1 & 31.8 & 35.9 & 36.5\\
	Lexical Variation (World Knowledge) & 8.3 & 4.7 & 9.9 & 20.3 & 18.8\\
	Syntactic Variation & 62.5 & 53.6 & 50.5 & 53.1 & 46.4\\
	Multiple Sentence Reasoning & 8.9 & 8.3 & 16.7 & 12.0 & 16.7\\
	Ambiguous & 1.6 & 3.6 & 1.6 & 1.6 & 1.0\\
\bottomrule
\end{tabular}

    \label{table:reasoning_required}
\end{table*}

\subsection{Are Models Trained with More Data More Robust to Natural
Distribution Shifts?}
\begin{figure*}[ht!]
\centering
    \begin{subfigure}{\textwidth}
        \centering
        \includegraphics[width=\linewidth]{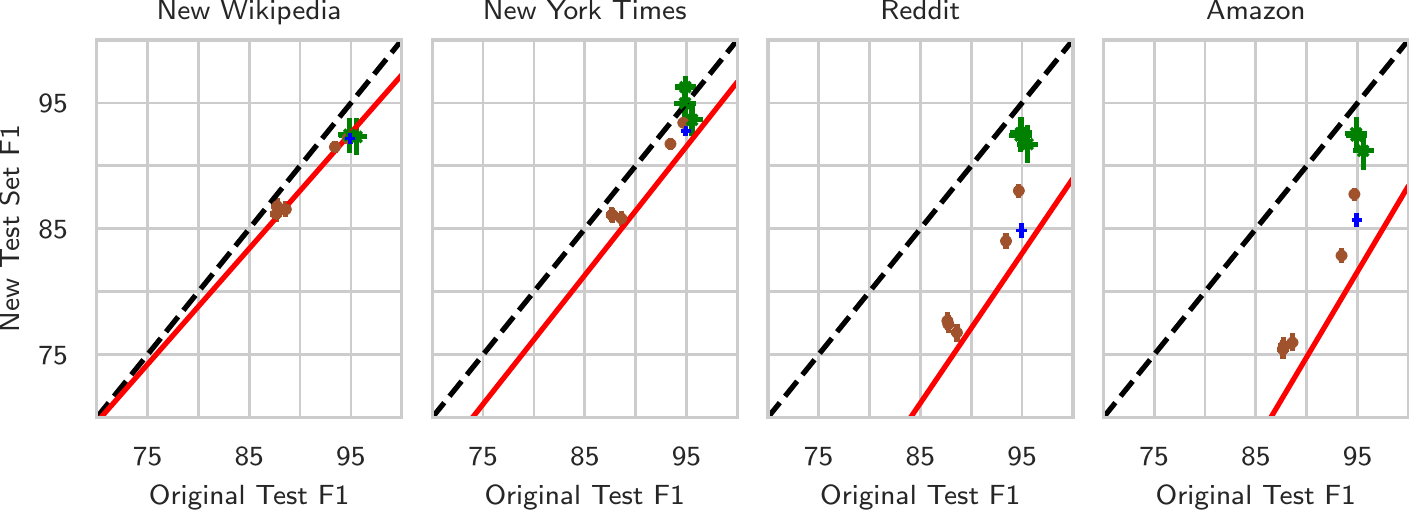}
    \end{subfigure} \\
    \vspace{0.05cm}
    \begin{subfigure}{\textwidth}
        \centering
        \includegraphics[width=0.9\linewidth]{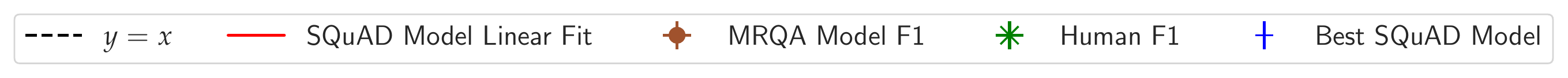}
    \end{subfigure} \\
    \caption{
        Model from the MRQA Shared Task 2019, trained on 5 datasets beyond
        \squad, and human F1 scores on the original \squad test set and each of
        our new test sets. The error bars are 95\% Student's t-confidence
        intervals.  Although the MRQA models still lag human performance and
        robustness across datasets, these models, particularly those with high
        F1 scores on the original \squad, exhibit increased robustness and
        generalization across each of the datasets compared to models that are
        only trained on \squad.}
\label{fig:mrqa_scatterplot}
\end{figure*}

High performance on our new datasets requires models to generalize to data
distributions that may be different from those on which they were trained.  Our
primary evaluation only concerns the robustness of \squad models, and a natural
follow-up question is whether models trained on more data, or explicitly trained for
out-of-distribution question-answering, perform better on our new test sets. 

To test this claim, we evaluated a collection of models from the Machine Reading
for Question Answering (MRQA) 2019 Shared Task on
Generalization~\citep{fisch2019mrqa}. In the shared task, models were trained on
6 question-answering datasets, including \squad v1.1, and then evaluated on 12
held-out datasets. The datasets simultaneously differed not just in the passage
distribution, as in our experiments, but also in confounders like the data
collection procedure, the question distribution, and the relationship
between questions and passages.

In Figure~\ref{fig:mrqa_scatterplot}, we plot the F1 scores of MRQA models on
the \squad v1.1 dataset against the F1 scores on each of our new test
sets, along with the linear fits from Figure~\ref{fig:main_figure}. On the
Reddit and Amazon test sets, the best MRQA model in our testbed, Delphi
\citep{longpre2019exploration}, achieves higher F1 scores than any \squad model
and is substantially above the linear fit. However, many of the models trained
on more data exhibit little to no improved robustness. In addition, all of the
models are still substantially below the human F1 scores and robustness.  See
Appendix~\ref{app:results_tables} for the full results table.

\section{Discussion}
\label{sec:discussion}
Despite years of test set reuse, we find no evidence of adaptive overfitting on
\squad. Our findings demonstrate that natural language processing benchmarks like
\squad continue to support progress much longer than than reasoning from first
principles might have suggested.

While \squad models generalize well to new examples from the same
distribution, results on our new test sets also show that
robustness to distribution shift remains a challenge.  On each of our new test
sets, a strong human baseline is largely unchanged, but \squad models suffer non-trivial
and nearly uniform performance drops. While question answering models have made
substantial progress on \squad, there has been less progress towards
closing the robustness gap under non-adversarial distribution shifts. This
highlights the need to move beyond model evaluation in the standard, i.i.d.\
setting, and to explicitly incorporate distribution shifts into evaluation. We hope
our new test sets offer a helpful starting point.

There are multiple promising avenues for future work. One direction is
constructing metrics for comparing datasets that can explain the performance
differences we observe. Why do models perform so well on New York Times, but
experience much larger drops on Reddit and Amazon? Stratifying our datasets
using common criteria like answer type or reasoning required appears
insufficient to answer this question.  Another important direction is to better
understand the interplay between additional data and model robustness.  Some of
the models from the MRQA challenge, e.g.,  Delphi~\citep{longpre2019exploration},
benefit substantially from training with additional data, while other models remain
near the same linear trend line as the \squad models. From both empirical and
theoretical perspectives, it would be interesting to better understand when and
why training with additional data improves robustness, and to offer
concrete guidance on how to collect and use additional data to improve robustness
to distribution shifts.

\section*{Acknowledgments}
We thank Pranav Rajpurkar, Robin Jia, and Percy Liang for providing us with the
original SQuAD data generation pipeline and answering our many questions about
the SQuAD dataset. We thank Nelson Liu for generously providing many of the
SQuAD models we evaluated, substantially increasing the size of our testbed.  We
also thank the Codalab team for supporting our model evaluation efforts. This
research was generously supported in part by the National Science Foundation
Graduate Research Fellowship Program under Grant No.  DGE 1752814 ABC, an Amazon
AWS AI Research Award, and a gift from Microsoft Research.

\bibliographystyle{plainnat}
\bibliography{squad}

\clearpage
\newpage
\onecolumn
\appendix

\etocdepthtag.toc{mtappendix}
\vspace{-.5cm}

\etocsettagdepth{mtsection}{none}
\etocsettagdepth{mtappendix}{subsubsection}
\tableofcontents

\section{Evaluation Metrics} \label{app:metrics}
In this section, we formally define the evaluation metrics used throughout our
experiments. Let $(p, q, (a^1, \dots, a^n)$ denote a passage $p$, a question
$q$, and a set of $n$ answers $(a^1, \dots, a^n)$. Let $S$ denote the sampled dataset, let $f$
denote some model, and $f(p, q) = \hat{a}$ be its predicted answer.

\paragraph{F1 Score.} F1 measures the average overlap between the prediction and the
ground-truth answer. Given answer $a$ and prediction $\hat{a}$, consider $a$ and
$\hat{a}$ as bags of words (sets), and let $v(a, \hat{a})$ be their associated
F1 score, i.e. the harmonic mean of precision and recall between the two sets.
Then,

\begin{align*}
    \mbox{F1}(f) = \frac{1}{|S|} \sum_{(p, q, (a^1, \dots, a^n)) \in S}
    \max_{i = 1, \dots, n} v(a^i, f(p, q)).
\end{align*}

\paragraph{Exact match.} Exact match measures the percentage of predictions that
exactly match any one of the ground truth answers.
\begin{align*}
    \mbox{ExactMatch}(f) = \frac{1}{|S|} \sum_{(p, q, (a^1, \dots, a^n)) \in S}
    \max_{i = 1, \dots, n} \mathbbm{1}\{f(p, q) = a^i\}.
\end{align*}

All of our results are reported using the evaluation script provided
by~\citet{rajpurkar2016squad}, which ignores punctuation and the articles ``a'',
``an'', and ``the'' when computing the above metrics.

\section{Comparing Natural and Adversarial Distribution Shift}
\label{app:adversarial}
To contrast natural and adversarial distribution shifts, we evaluated all of the
models in our testbed against the adversarial attacks described
in~\citet{jia2017adversarial} on the original SQuAD v1.1 dataset.

\begin{figure}[ht!]
\centering
    \begin{subfigure}{\textwidth}
        \centering
        \includegraphics[width=\linewidth]{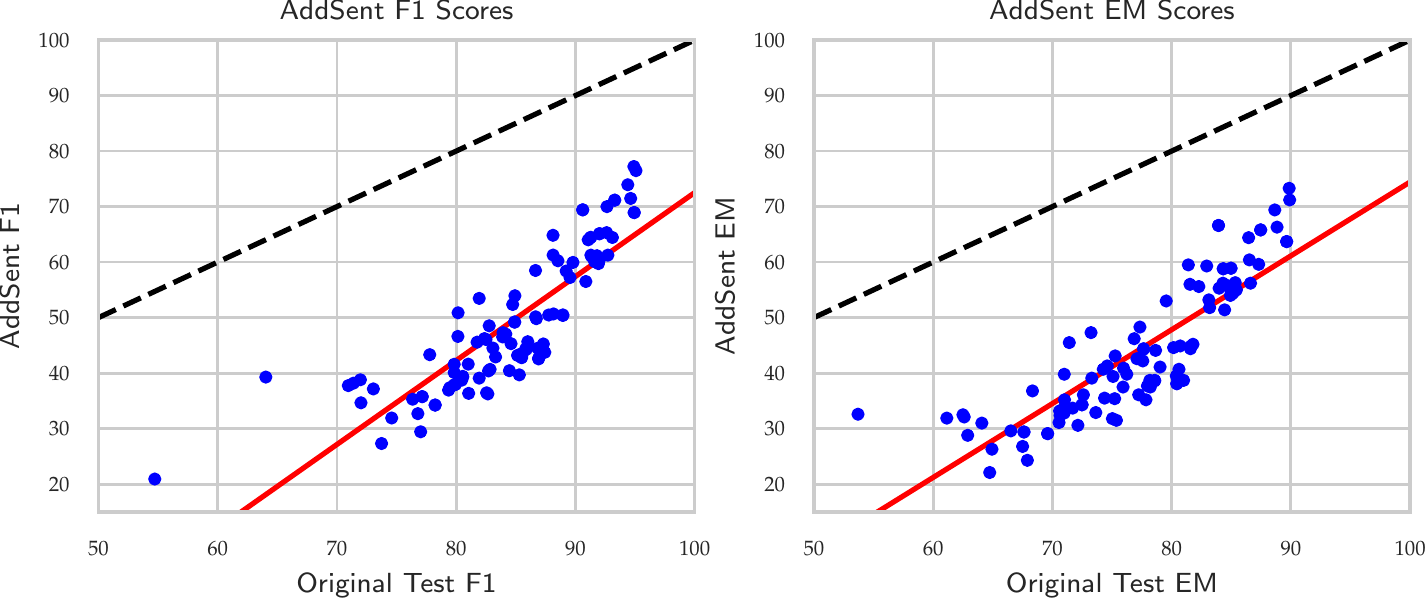}
    \end{subfigure} \\
    \vspace{0.05cm}
    \begin{subfigure}{\textwidth}
        \centering
        \includegraphics[width=0.3\linewidth]{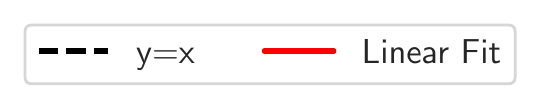}
    \end{subfigure} \\
    \caption{Comparison of F1 and EM scores on the original \squad test set
    versus the \emph{adversarial} {\tt AddSent} attack
    from~\citet{jia2017adversarial}. The models exhibit substantially more
    variability around the linear trend line compared to natural distribution
    shifts. For F1 scores, the slope of the linear fit is 1.51, for EM scores,
    the slope is 1.33. Similarly, the $R^2$ statistics are $0.73$ and $0.74$,
    respectively.}
    \label{fig:adv_addsent}
\end{figure}

\paragraph{AddSent.}
In the {\tt AddSent} attack, for every passage, question, and answer pair $(p,
q, a)$, \citet{jia2017adversarial} procedurally generate up to five new
sentences to append to the passage $p$ that do not contradict the correct
answer. Each of the sentences are generated to be similar to the correct answer,
and ungrammatical or contradictory sentences are removed by crowdworkers.
This results in a set of new examples $(\tilde{p}_1, q, a), \dots, (\tilde{p}_5,
q, a)$ for each original example. The adversary evaluates the model $f$ on each
of the 5 examples and picks the one that gives the lowest score, $\min_{i=1,
\dots, 5} s(f(\tilde{p}_i, q), a)$, where $s$ is the scoring function (exact
match or F1).  In Figure~\ref{fig:adv_addsent}, we compare F1 and EM scores
on the original \squad v1.1 test set with F1 and EM scores against the adversarial
{\tt AddSent} attack. 

\begin{figure}[ht!]
\centering
    \begin{subfigure}{\textwidth}
        \centering
        \includegraphics[width=\linewidth]{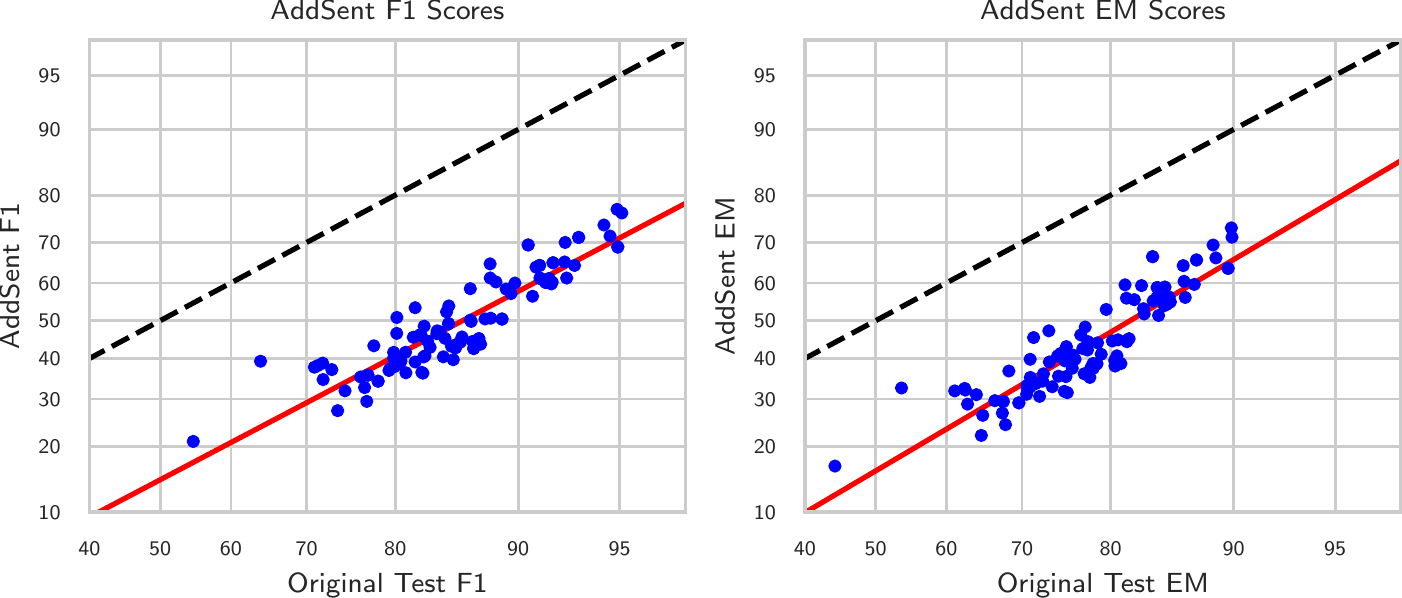}
    \end{subfigure} \\
    \vspace{0.05cm}
    \begin{subfigure}{\textwidth}
        \centering
        \includegraphics[width=0.3\linewidth]{figures/adv_addsent_legend}
    \end{subfigure} \\
    \caption{Comparison of F1 and EM scores on the original \squad test set versus the
    \emph{adversarial} {\tt AddSent} attack from~\citet{jia2017adversarial}
    with \emph{probit} scaling. For F1 scores, the slope of the linear fit is
    0.99, and for EM, the slopes  is 1.11. In the probit domain, the $R^2$
    statistics are $0.82$ and $0.81$, respectively.}
    \label{fig:probit_adv_addsent}
\end{figure}

Similar to the natural distribution shift examples, we observe the relationship
between the original test F1 scores and the adversarial F1 test scores broadly
follow a linear trend. However, the linear fit is not as good compared to the
natural distribution shifts. There is more variability in model performance
around the trend line, and this is reflected in lower a $R^2$ statistic, e.g.
$0.72$ for AddSent F1, compared to 0.99, 0.97, 0.91, and 0.89 for the New
Wikipedia, New York Times, Reddit, and Amazon test sets, respectively. As with
the natural distribution shift datasets, the linear fit is better in the probit
domain, which we visualize in Figure~\ref{fig:probit_adv_addsent}. However, the
$R^2$ statistic is still smaller than the corresponding statistics for our
distribution shift datasets in the probit domain: $0.82$ compared to $0.99,
0.96, 0.94$, and $0.94$, for New Wikipedia, New York Times, Reddit, and Amazon,
respectively.

\begin{figure}[ht!]
\centering
    \begin{subfigure}{\linewidth}
        \centering
        \includegraphics[width=\linewidth]{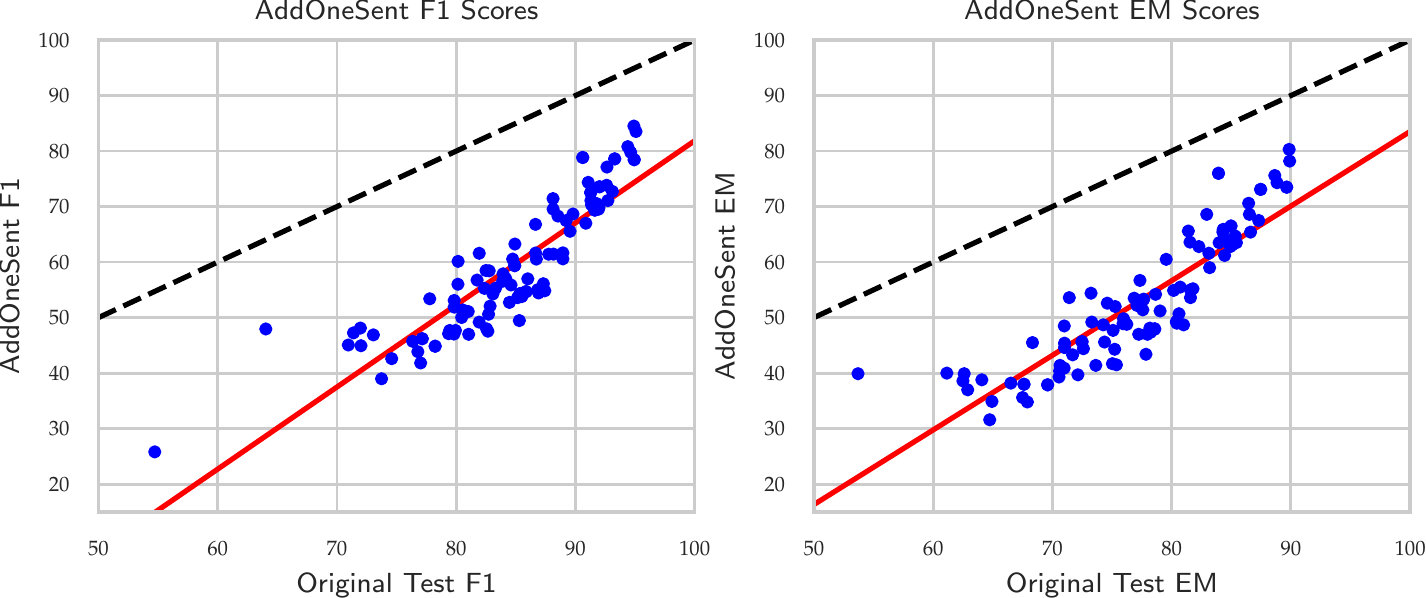}
    \end{subfigure} \\
    \vspace{0.05cm}
    \begin{subfigure}{\linewidth}
        \centering
        \includegraphics[width=0.35\linewidth]{figures/adv_addsent_legend}
    \end{subfigure} \\
    \caption{Comparison of F1 and EM scores on the original \squad test set versus the
    \emph{adversarial} {\tt AddOneSent} attack from~\citet{jia2017adversarial}.
    We observe similar phenomenon as with {\tt AddSent}. Model performance
    broadly follows a linear trend, with more variability around the trend line
    than with our natural distribution test sets.  For F1 scores, the slope of
    the linear fit is $1.48$, and for EM, the slopes is $1.34$. The $R^2$
    statistics are $0.79$ and $0.80$, respectively.}
    \label{fig:adv_addonesent}
\end{figure}

\paragraph{AddOneSent.}
The {\tt AddOneSent} attack similar to the {\tt AddSent} attack. However, rather
than take the worst of the 5 altered passages, it randomly selects one of the
five on which to evaluate the model. In Figure~\ref{fig:adv_addonesent}, we
compare F1 and EM scores on the original \squad v1.1 test set with F1 and EM
scores against the adversarial {\tt AddSent} attack. 
Since this attack does not require model access or evaluations, it is closer in
spirit to the natural distribution shifts we consider. We observe much
the same phenomenon as we see with {\tt AddSent}. Model performance broadly
follows a linear trend, and there is more variability around the linear trend
line than in our natural distribution shift datasets.

\FloatBarrier

\section{Additional Analysis and Results}
\label{app:bonus_results}

In this appendix, we present additional results and analysis to better
understand our distribution shift experiments.

\subsection{Exact Match Scatterplots}
\label{app:scatterplots}
\begin{figure*}[ht!]
\centering
    \begin{subfigure}{\textwidth}
        \centering
        \includegraphics[width=\linewidth]{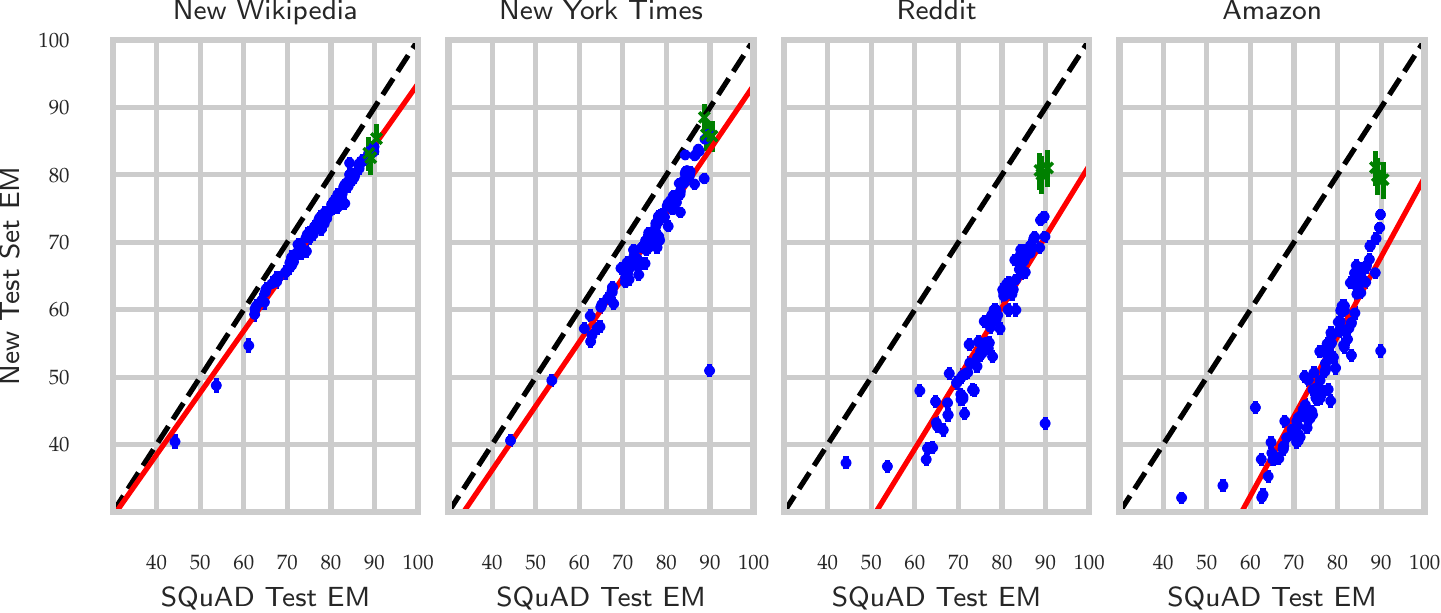}
    \end{subfigure} \\
    \vspace{0.05cm}
    \begin{subfigure}{\textwidth}
        \centering
        \includegraphics[width=0.65\linewidth]{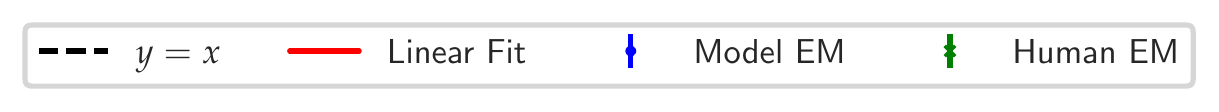}
    \end{subfigure} \\
    \caption{
        Model and human EM scores on the original SQuAD test set compared to our
        new test sets (shown with 95\% Clopper-Pearson confidence intervals).
        The slopes of the linear fits are 0.92, 0.95, 1.05, and 1.18,
        respectively. The $R^2$ statistics are $0.99, 0.83, 0.82$, and $0.85$,
        respectively.}
\label{fig:em_scatterplot}
\end{figure*}

Similar to Figure~\ref{fig:main_figure} in Section~\ref{sec:intro}, we
compare the EM scores of all models in our testbed on the \squad v1.1 test set
versus the EM scores of all models on each of the new test sets. The results
are shown in Figure~\ref{fig:em_scatterplot}. In each case,
we observe a more pronounced drop than the F1 scores with average drops of 4.6,
5.75, 20.0, and 24.8 for each of the new Wikipedia, New York Times, Reddit, and
Amazon datasets, respectively. However, the primary trends are the same. In
particular, we observe little evidence of overfitting on Wikipedia (the linear
model nicely describes the data), and we observe a similar ranking of magnitudes
of the drop on each of the other three datasets--- New York Times exhibits a small
drop, followed by larger drops on Reddit and Amazon.

\FloatBarrier

\subsection{Linear Fits in the Probit Domain}
\label{app:probit_fits}
\begin{figure*}[ht!]
\centering
    \begin{subfigure}{\textwidth}
        \centering
        \includegraphics[width=\linewidth]{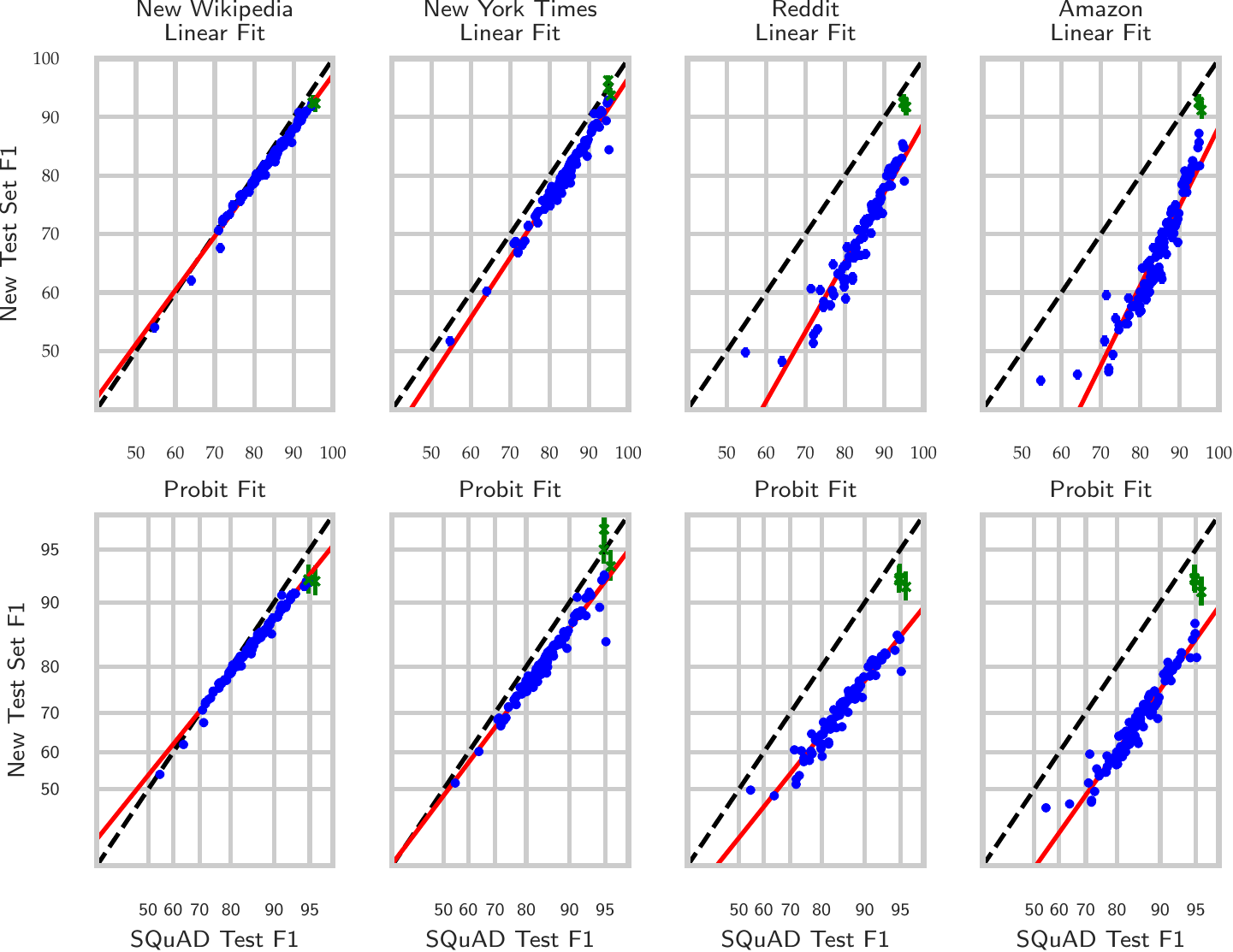}
    \end{subfigure} \\
    \vspace{0.05cm}
    \begin{subfigure}{\textwidth}
        \centering
        \includegraphics[width=0.65\linewidth]{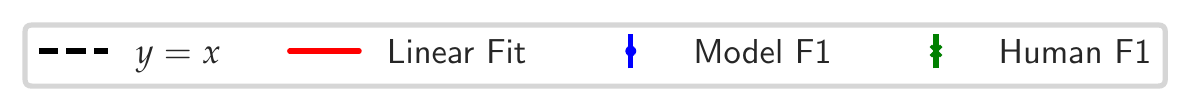}
    \end{subfigure} \\
    \caption{
        Comparison between linear and probit axis scaling for model and human F1
        scores on the original SQuAD test and each of our new test sets. For
        linear axis scaling, the slopes of the linear fit are $0.92$, $1.02$,
        $1.19$, and $1.36$, respectively, and the $R^2$ statistics are $0.99,
        0.97, 0.91, 0.89$, respectively.  Under probit axis scaling, the slopes
        of the linear fit are $0.83$, $0.89$, $0.84$, and $0.95$, respectively,
        and the $R^2$ statistics are $0.99, 0.96, 0.94, 0.94$, respectively.}
    \label{fig:probit_f1}
\end{figure*}

\begin{figure*}[ht!]
\centering
    \begin{subfigure}{\textwidth}
        \centering
        \includegraphics[width=\linewidth]{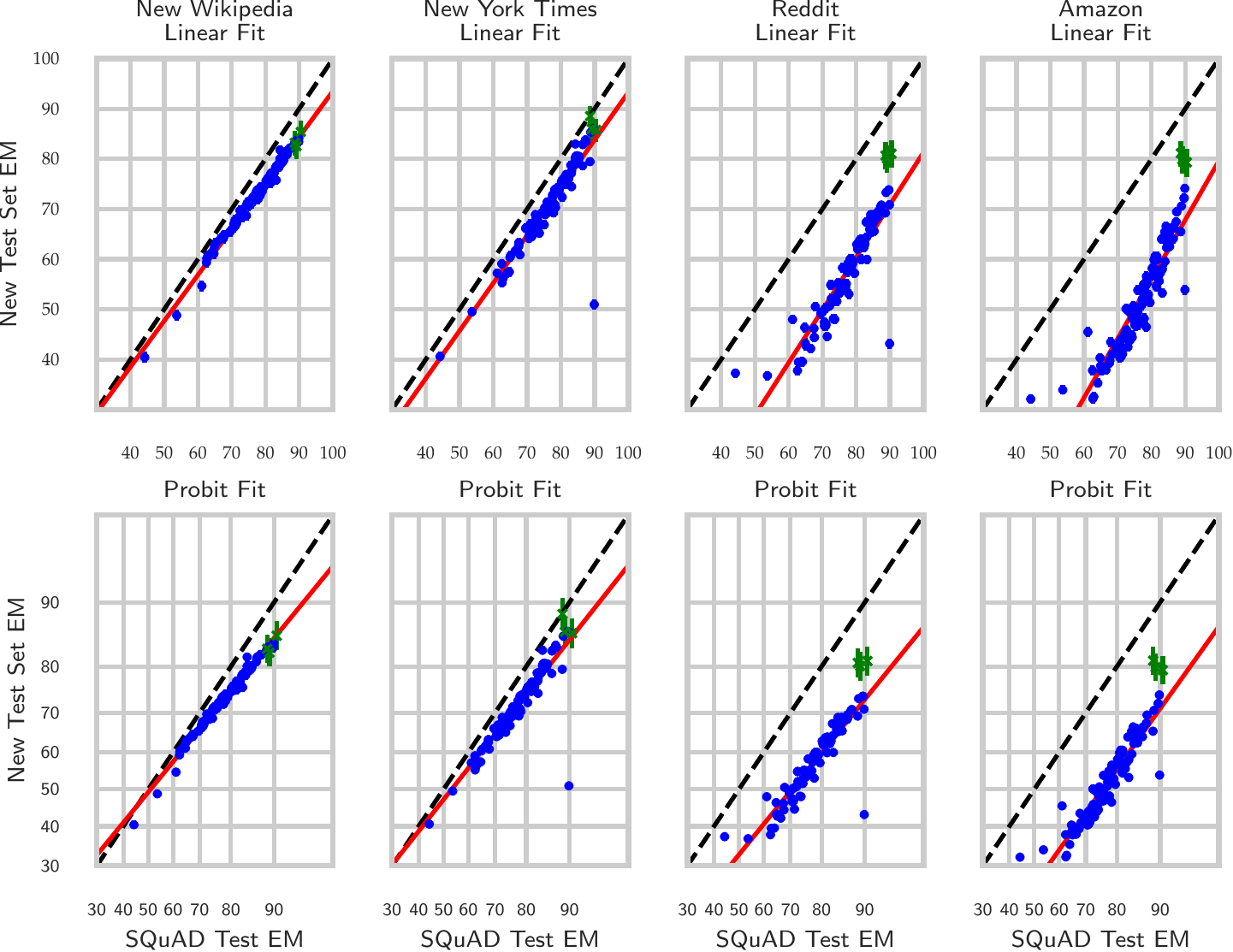}
    \end{subfigure} \\
    \vspace{0.05cm}
    \begin{subfigure}{\textwidth}
        \centering
        \includegraphics[width=0.65\linewidth]{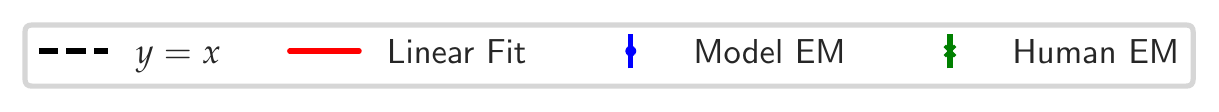}
    \end{subfigure} \\
    \caption{
        Comparison between linear and probit axis scaling for model and human EM
        scores on the original SQuAD test and each of our new test sets. Under
        linear axis scaling, the slopes of the linear fit are $0.92$, $0.95$,
        $1.05$, and $1.18$, respectively. The $R^2$ statistics are $0.99, 0.83,
        0.82$, and $0.85$, respectively. Under probit scaling, the slopes of the
        linear fit are $0.82$, $0.85$, $0.83$, and $0.94$, respectively. The
        $R^2$ statistics are $0.99, 0.82, 0.83$, and $0.88$, respectively.}
    \label{fig:probit_em}
\end{figure*}

In many cases, a linear model of F1 or EM scores is not a good fit when the
scores span a wide range. In these cases, we find that a probit model describes the
data better. In the main text, Figure~\ref{fig:comparison_probit} shows the F1
scores for the Amazon dataset on both the linear scale used throughout the data
and a probit scale obtained by transforming all of the F1 scores with the
inverse Gaussian CDF. We observe a better linear fit for our data.
Figures~\ref{fig:probit_f1} and Figures~\ref{fig:probit_em} show similar probit
models for each of our new datasets.

\FloatBarrier

\subsection{Does Annotator Agreement Correlate with Performance Drops?}
\label{app:label_agreement}

\begin{figure*}[ht!]
\centering
    \begin{subfigure}{\textwidth}
        \centering
        \includegraphics[width=\linewidth]{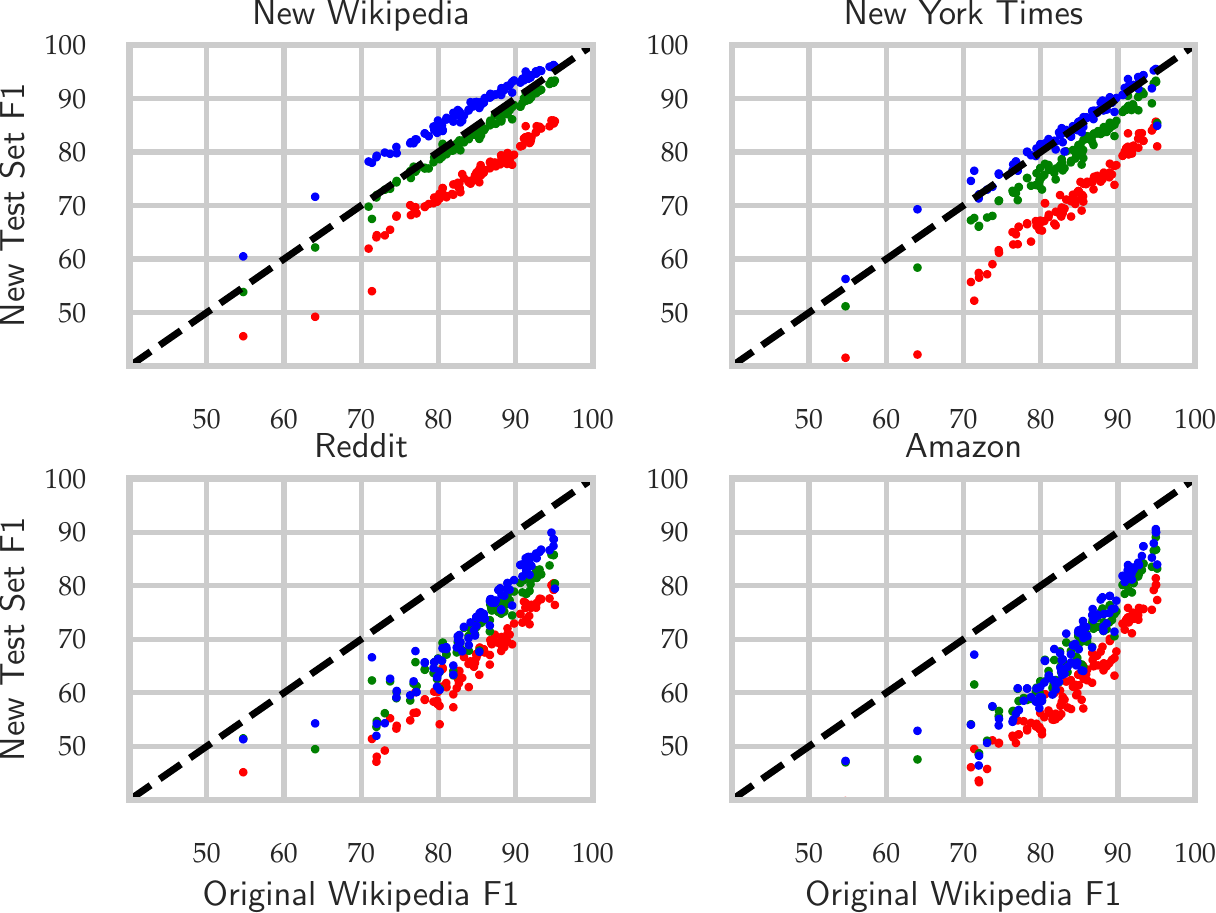}
    \end{subfigure} \\
    \vspace{0.1cm}
    \begin{subfigure}{\textwidth}
        \centering
        \includegraphics{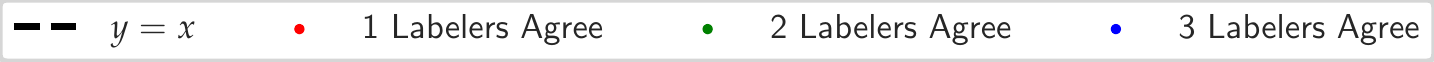}
    \end{subfigure} \\
    \caption{
        Model and human F1 scores on the original \squad v1.1 test set compared
        to our new test sets, stratified by the agreement between the answers
        given by the labellers, e.g. if three labellers agree, then three labellers
        provided identical (up to text normalization) answers to the question.
        Each point corresponds to a model evaluation. Label agreement roughly
        corresponds to question difficulty (and ambiguity). For clear and simple
        questions, all of the labellers typically agree. For more subtle or
        potentially ambiguous questions, the labeller's answers are more varied and
        tend to disagree more often. Across each dataset, when
        the questions are easier or less ambiguous (as measured by higher labeller
        agreement), the models experience proportionally smaller drops on the
        new dataset.
    }
\label{fig:annotator_agreement}
\end{figure*}
\FloatBarrier

\subsection{Do Shifts in Answer Category Distributions Predict Performance Drops?}
\label{app:answer_shifts}

\begin{figure*}[ht!]
    \centering
    \includegraphics[width=\linewidth]{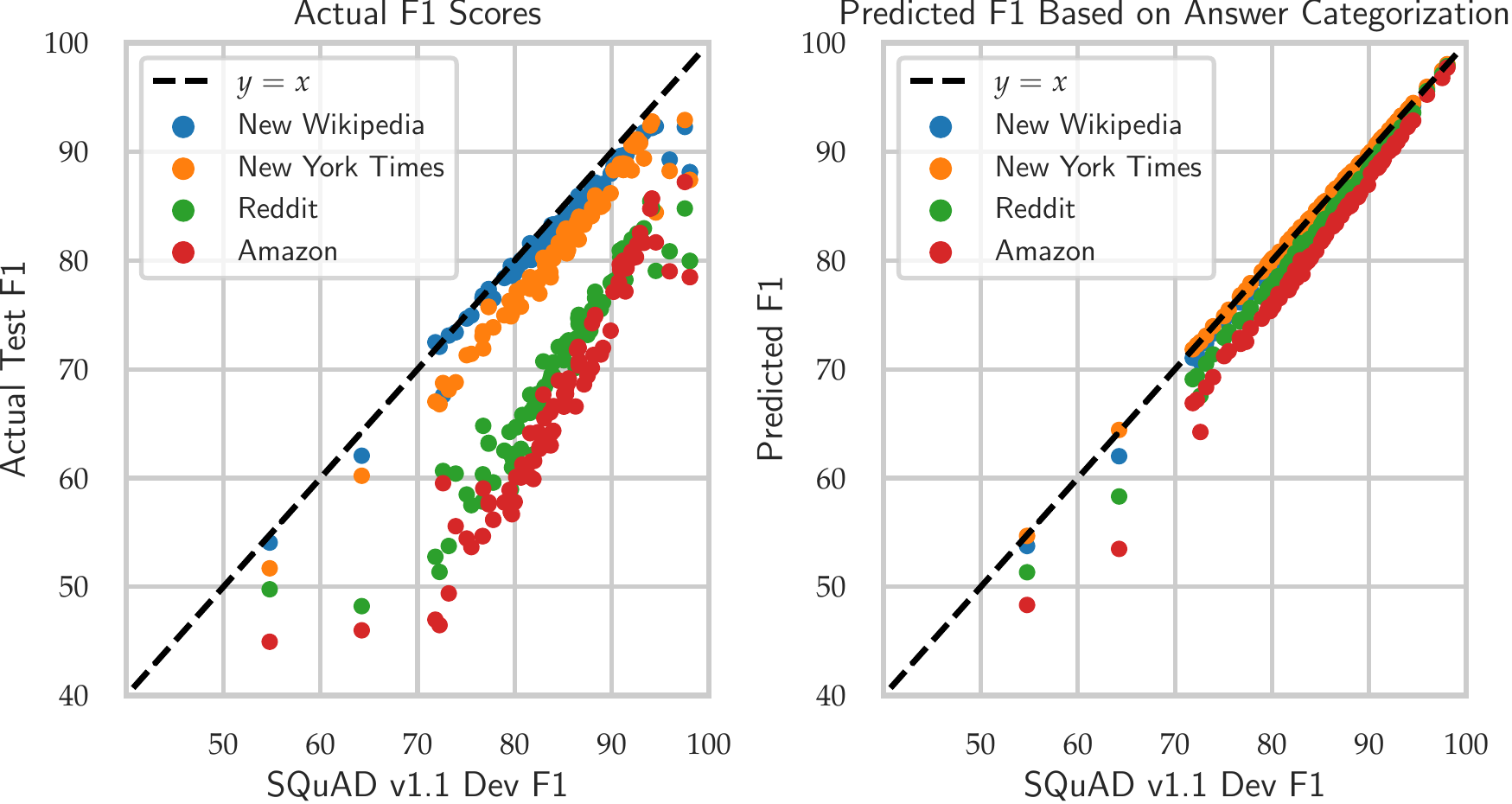}
    \caption{Changes in answer type distributions introduced in
    Section~\ref{sec:analysis} explain little of the observed
    performance differences across our new datasets. For each model, we compute the F1
    score on each of the answer types on the \squad v1.1 dev set, and then we
    predict the F1 score on the new test set by reweighing these F1 scores based
    on the frequency of answer types in the new test set. Concretely, if \squad
    v1.1 was 50\% {\tt NP} answers and 50\% {\tt Places} answers, and a model has
    average F1 scores of 100 for {\tt NP} and 75 for {\tt Places}, then if a
    new dataset had 30\% {\tt NP} answers and 70\% {\tt Places} answers, the
    predicted F1 score would be 82.5 (versus 87.5 for the original). The $y=x$
    line represents the trivial model that predicts the same F1 score on the new
    test sets as the original. For each of the distribution
    shift datasets, predictions based on answer category shifts are exceedingly
    optimistic and explain little of the observed drops. For instance, on the
    Reddit dataset, answer category shifts suggest models would lose, on
    average, 2-3 F1 points. However, the average observed shift is \redditdrop
    F1 points.
    }
\label{fig:answer_category_shift_model}
\end{figure*}
\FloatBarrier

\subsection{Do Shifts in Syntactic Divergence Predict Performance Drops?}
\label{app:syntax_shifts}

\begin{figure*}[ht!]
    \centering
    \includegraphics[width=\linewidth]{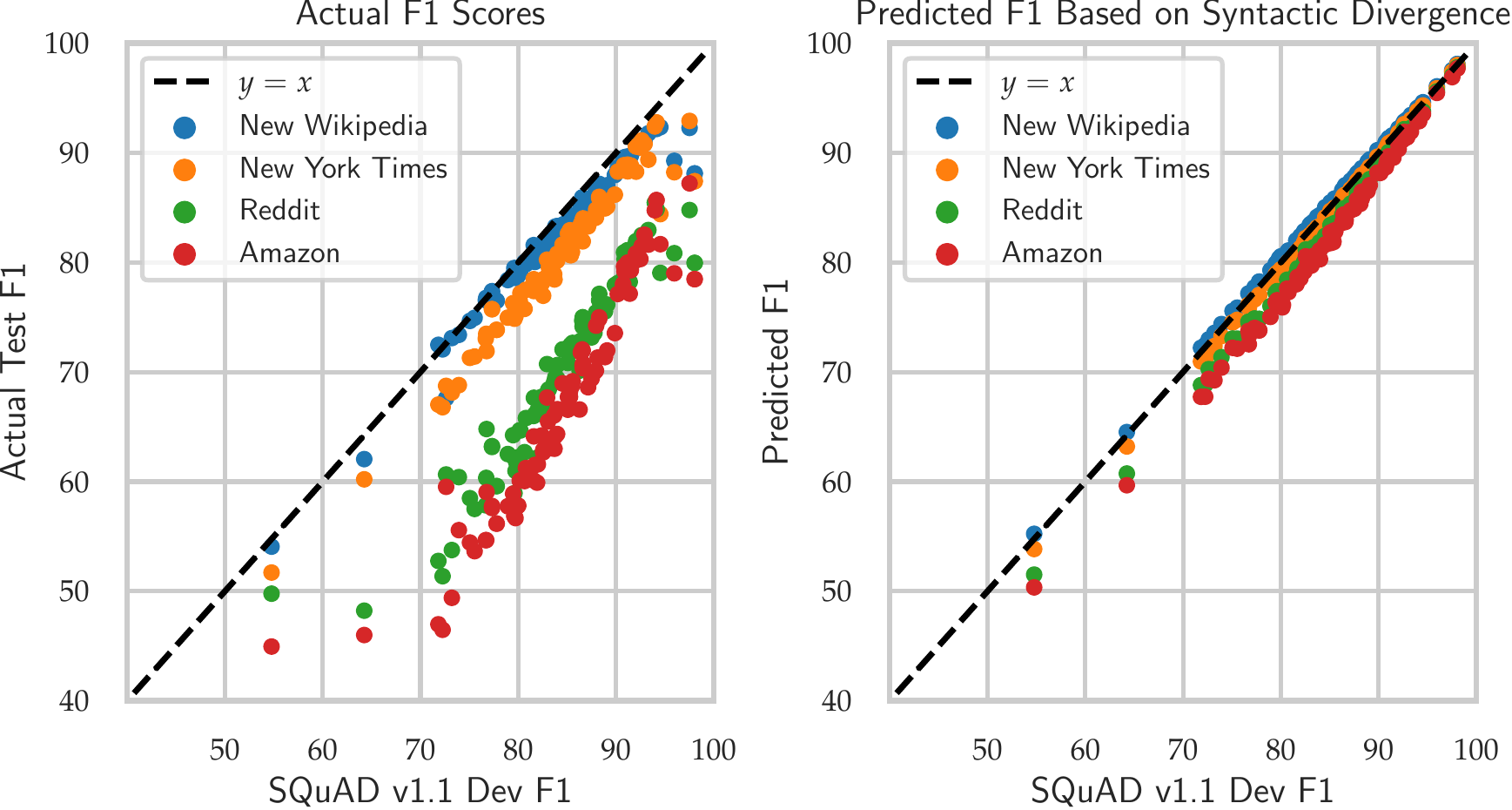}
    \caption{Changes in syntactic distributions introduced in
    Section~\ref{sec:analysis} explain only a small amount of the observed
    performance differences across our new datasets. As in the previous plot, 
    for each model, we compute the F1 score for each observed value of syntactic
    divergence on the \squad v1.1 dev set, and then we predict the F1 score on
    the new test set by reweighing these F1 scores based on the frequency of
    examples with a given syntactic divergence in the new test set. 
    For each of the distribution shift datasets, predictions based on answer
    category shifts are optimistic. For instance, on the Reddit dataset,
    syntactic divergence shifts suggest models would lose, on average, 1.9 F1
    points, while the average observed shift is \redditdrop F1 points.
    }
\label{fig:syntax_shift_model}
\end{figure*}
\FloatBarrier

\subsection{Do Shifts in Reasoning Required Distributions Predict Performance Drops?}
\label{app:reasoning_shifts}

\begin{figure*}[ht!]
    \centering
    \includegraphics[width=\linewidth]{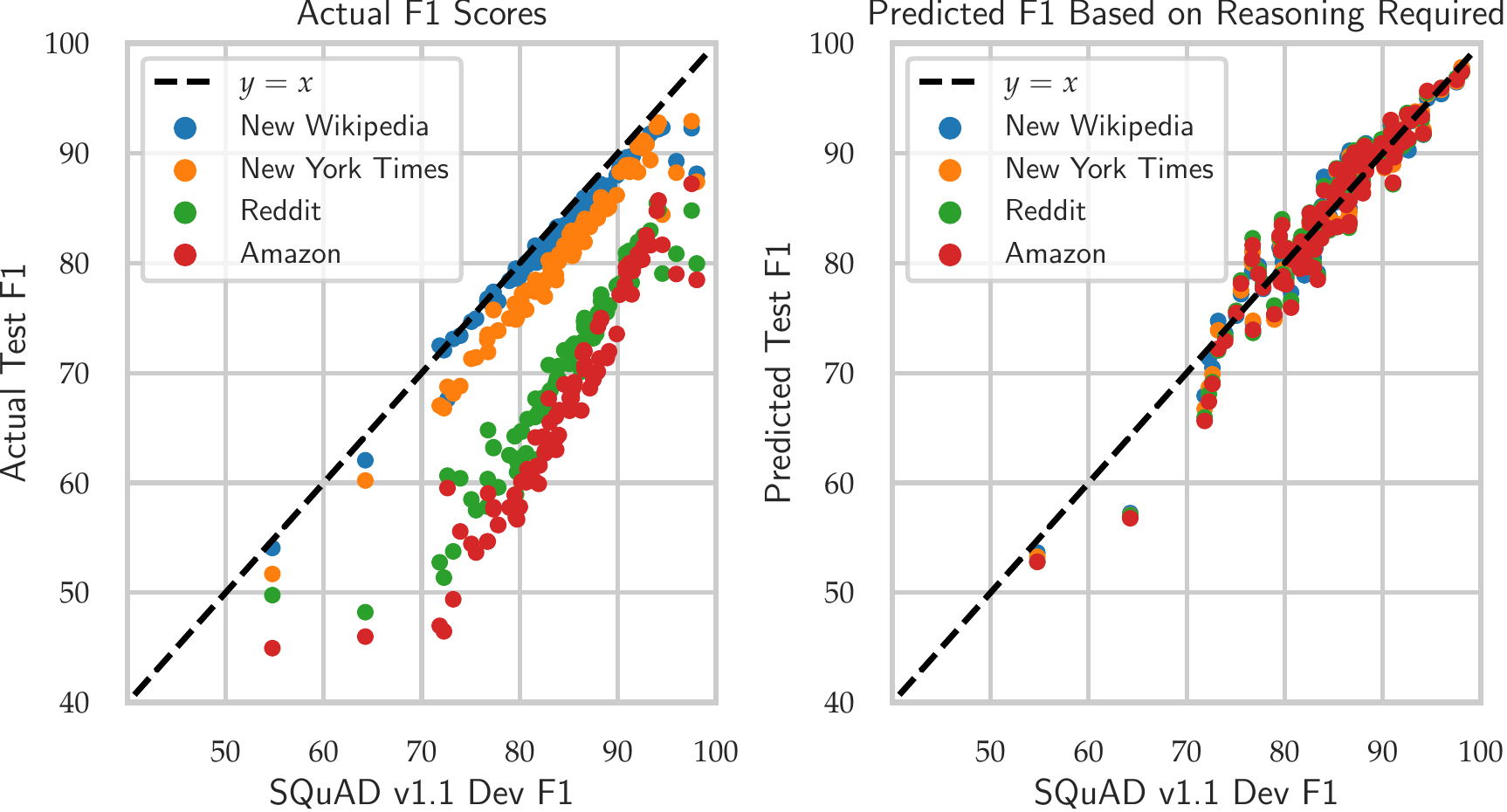}
    \caption{Changes in reasoning required distributions introduced in
    Section~\ref{sec:analysis} explain little of the observed
    performance differences across our new datasets. Similar to the previous
    plot, for each model, we compute the F1 score on each of the reasoning
    required categories on the \squad v1.1 dev set, and then we predict the F1 score
    on the new test set by reweighing these F1 scores based on the reasoning
    required distribution of the new test set. For each of the distribution
    shift datasets, predictions based on reasoning required shifts closely follow the 
    $y=x$ line corresponding to the trivial model that predicts the same F1
    score on the new test sets as the original.}
\label{fig:reasoning_required_shift_model}
\end{figure*}
\FloatBarrier

\subsection{Does Manual Data Curation Reduce Performance Drops?}
\label{app:data_cleaning}
\begin{figure*}[ht!]
    \centering
    \includegraphics[width=\linewidth]{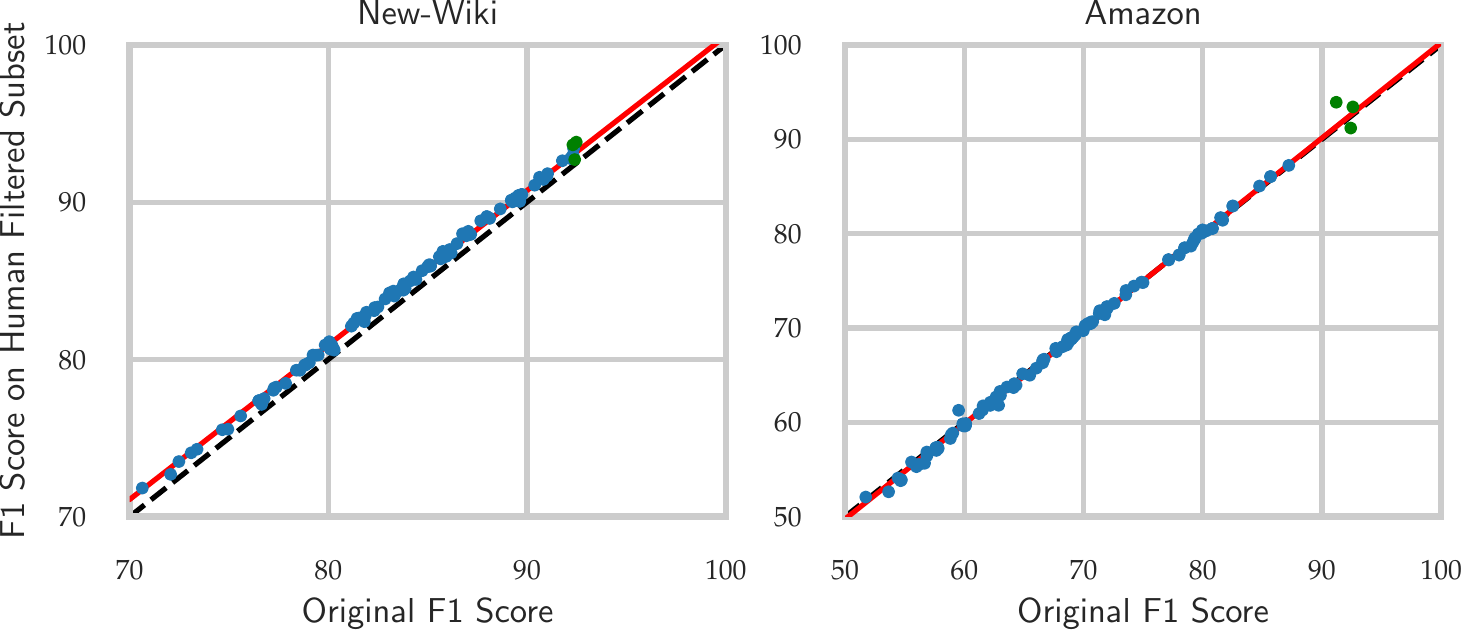}
    \caption{Comparison between model F1 scores on our New-Wikipedia and Amazon
    datasets and F1 scores on subsets of the datasets with additional human
    filtering to remove malformed, unanswerable, incorrect, ungrammatical
    questions and answers. To focus annotator effort on potentially bad
    questions, if all three MTurk annotators agreed on the answer, the
    question and answer were automatically marked as valid.  For the New
    Wikipedia dataset, we manually inspected an additional 1,894 questions,
    removed 85 questions, and removed answers for an
    444 questions.  For the Amazon dataset, we manually inspected an
    additional 1,839 questions, removed 46 questions, and
    removed answers for an 282 questions. This process resulted in
    human curated subsets of 5574 questions for the New Wikipedia dataset and
    6471 questions for the Amazon datasets. On the New Wikipedia dataset, models
    improve an average of 0.86 F1 points on this filtered dataset. For the
    filtered Amazon dataset, models slightly decreased their performance by 0.09
    F1 points on average. In both cases, the rank order of the models and the
    linear trend observed on the full datasets without additional human
    filtering is preserved.}
\label{fig:datacleaning_comparison}
\end{figure*}
\FloatBarrier

\section{Dataset collection details.}
In this section, we provide further details regarding our data collection
pipeline.

\subsection{Passage Length Statistics} \label{app:length_statistics}
We report statistics on various text length statistics. We split each paragraph into
individual sentences, words, and characters using spaCy~\citep{spacy2} and
compute histograms showing the passage sentence, word, and character length
distributions across each dataset.

\begin{figure}[ht!]
    \centering
    \includegraphics[width=0.8\textwidth]{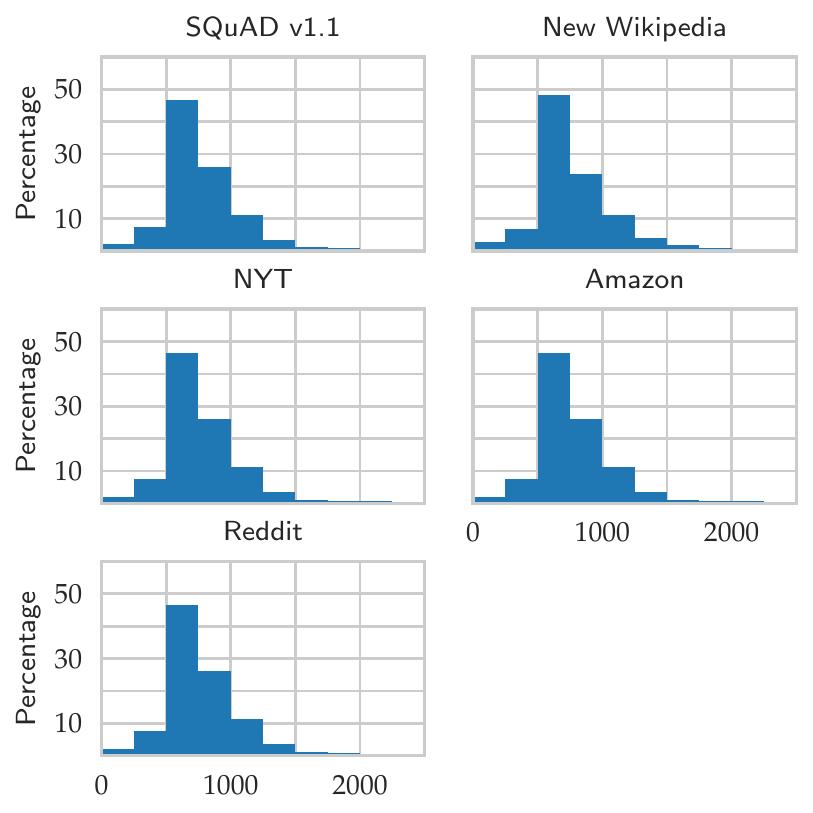}
    \caption{Histograms of the number of characters in each paragraph for the
    original \squad v1.1 development set and our new test sets. The histograms lengths
    match exactly since we sample in a way that ensures the character length
    will match for each new dataset.}
    \label{fig:char_paragraph_lengths}
\end{figure}

\begin{figure}[ht!]
    \center
    \includegraphics[width=0.8\linewidth]{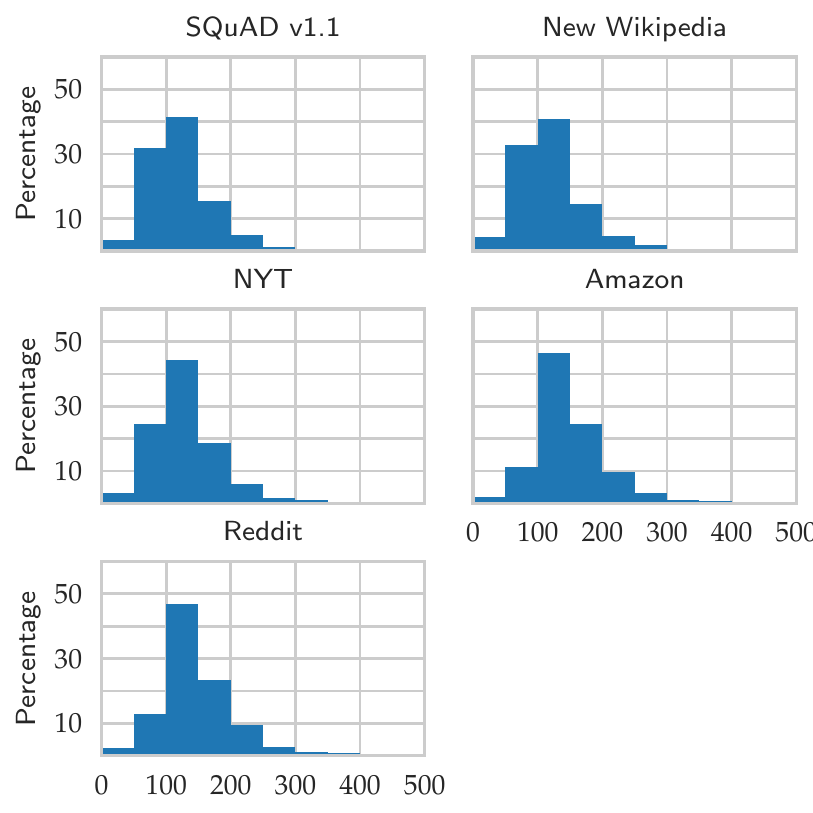}
    \caption{Histograms of the number of words in each paragraph for the
    original \squad v1.1 development set and our new test sets. The Wikipedia
    histograms match closely, while the Amazon and Reddit datasets' paragraphs
    have slightly more words. However, these differences do not explain the
    performance drops we observe, as Figure~\ref{fig:word_length_model}
    demonstrates.}
    \label{fig:word_paragraph_lengths}
\end{figure}

\begin{figure}[ht!]
    \center
    \includegraphics[width=0.8\linewidth]{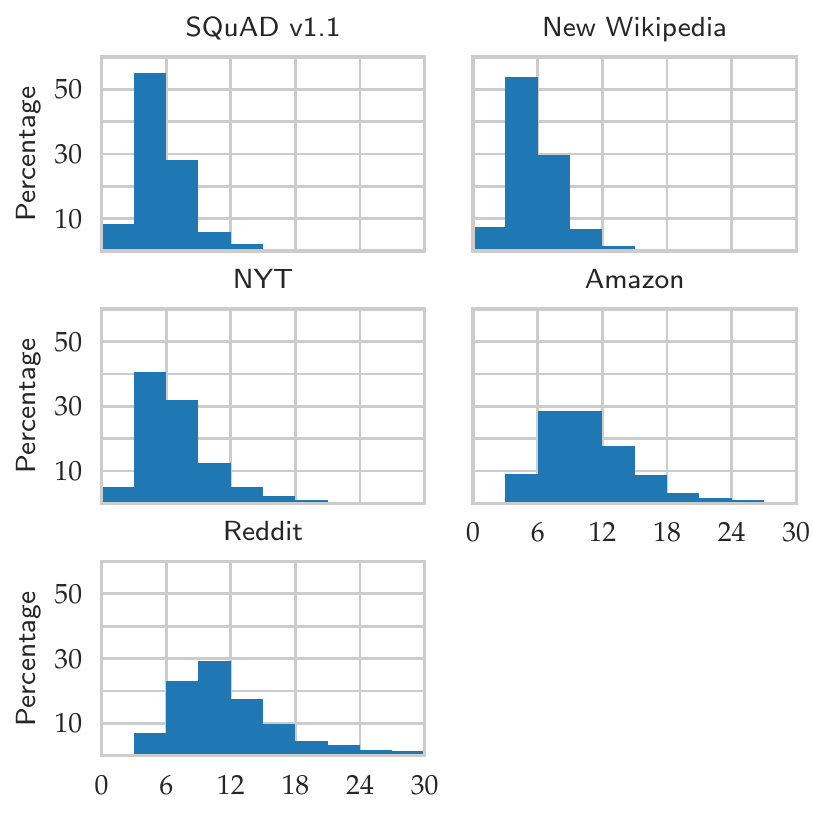}
    \caption{Histograms of the number of sentences in each paragraph for both the
    original and new datasets. The new Wikipedia dataset matches the \squad v1.1 dataset,
    while the other new test sets have a slightly longer tail. These slight
    difference in sentences per paragraph do not explain the performance drops
    we observe, as Figure~\ref{fig:sentence_length_model} demonstrates.}
    \label{fig:sentence_paragraph_lengths}
\end{figure}

\begin{figure}[ht!]
    \centering
    \includegraphics[width=\linewidth]{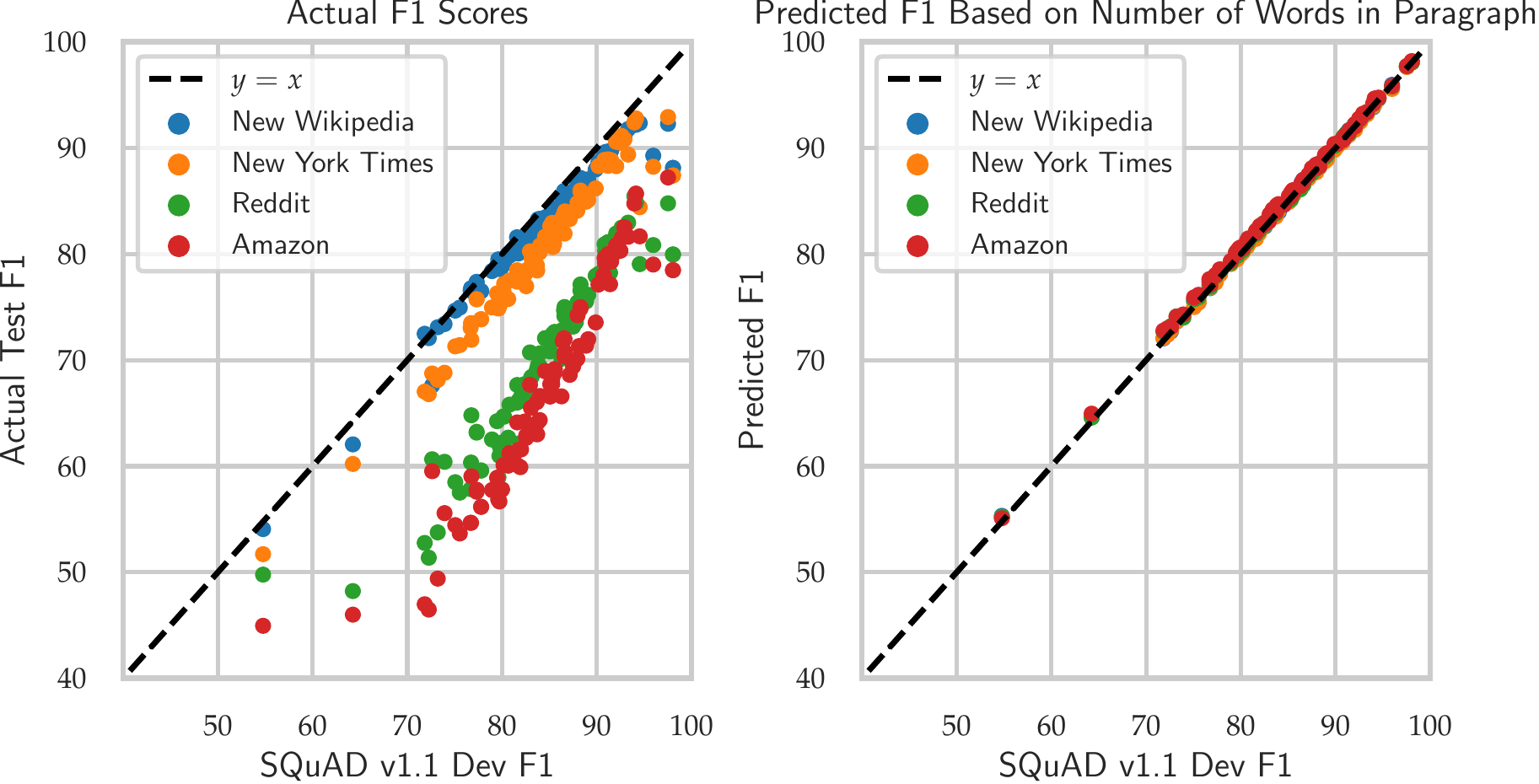}
    \caption{Changes in the distribution of words per paragraph across our new
    test sets do not explain the differences in F1 scores we observe.
    Concretely, we stratify the datasets by words per paragraph, and,
    for each model, we compute the F1 score for each bucket on the \squad v1.1
    development set. We then predict F1 scores on the new test set by reweighing
    these F1 scores based on the paragraph length distribution of the new test
    set.}
    \label{fig:word_length_model}
\end{figure}

\begin{figure}[ht!]
    \centering
    \includegraphics[width=\linewidth]{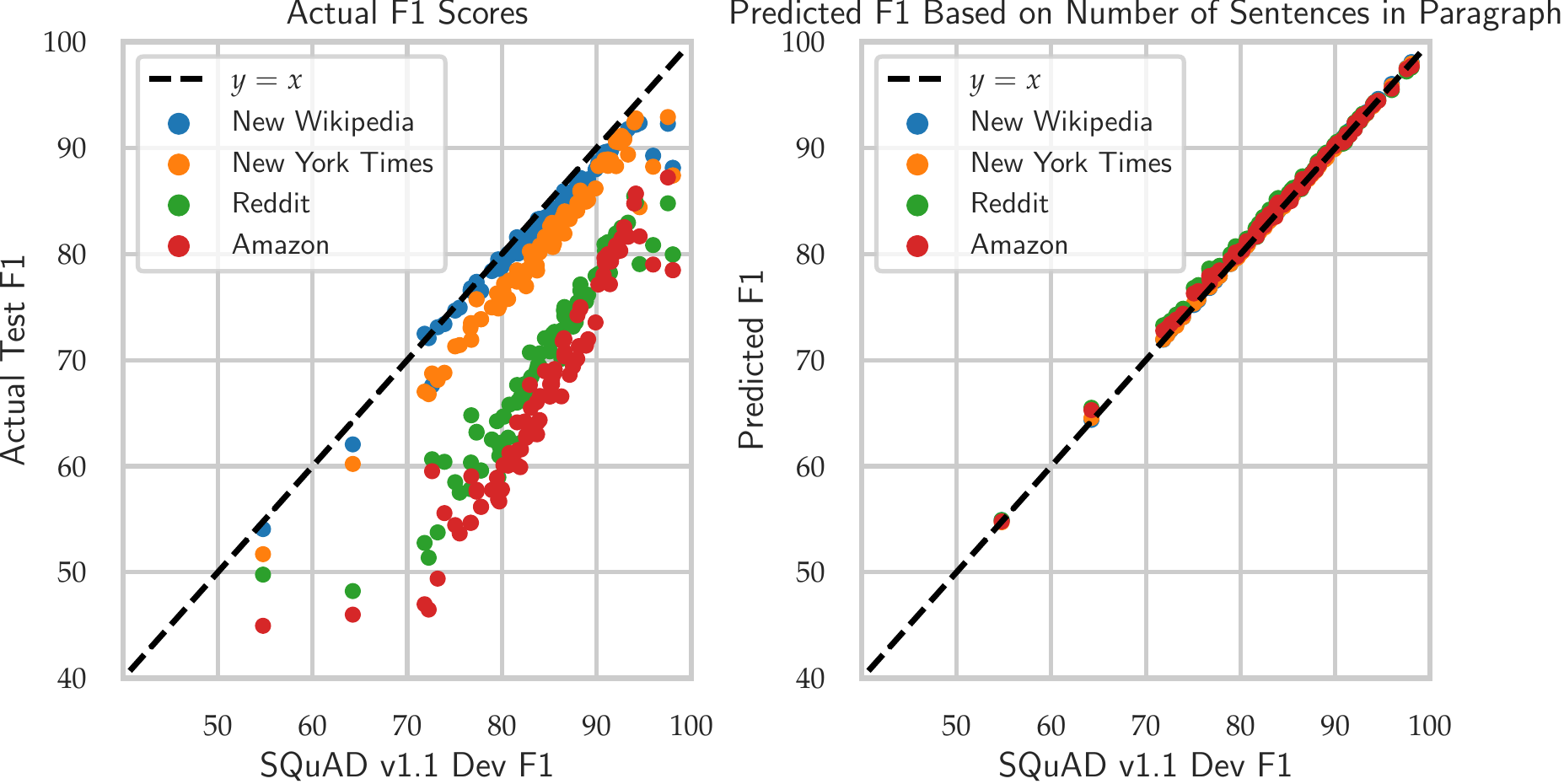}
    \caption{Changes in the distribution of sentences per paragraph across our new
    test sets do not explain the differences in F1 scores we observe.
    As in the previous plot, we stratify the datasets by sentence per paragraph,
    compute the F1 score for each bucket on the \squad v1.1
    development set, and then predict F1 scores on the new test set by reweighing
    these F1 scores based on the paragraph length distribution of the new test
    set.}
    \label{fig:sentence_length_model}
\end{figure}

Figures~\ref{fig:char_paragraph_lengths},~\ref{fig:word_paragraph_lengths},~and~\ref{fig:sentence_paragraph_lengths}
show the paragraph lengths in characters, words, and sentences across each
dataset. In
Figures~\ref{fig:word_length_model}~and~\ref{fig:sentence_length_model}, we show
the small differences in the distribution of passage length in terms of words or
sentences does not explain the observed performance drops.

\FloatBarrier


\subsection{MTurk Experiment and UI Examples}
\label{app:mturk_details}

\paragraph{Worker Details.}
Crowdworkers were required to have a 97\% HIT acceptance rate, a minimum of 1000
HITs, and be located in the United States or Canada.  Workers were asked to
spend four minutes per paragraph when asking questions and one minute per
question when answering questions. We paid workers \$9.60 per hour for the amount
of time required to complete each task, using an inflation rate of 6.52\%
between 2016 and 2019.

\begin{figure*}[ht!]
    \centering
    \includegraphics[width=\linewidth]{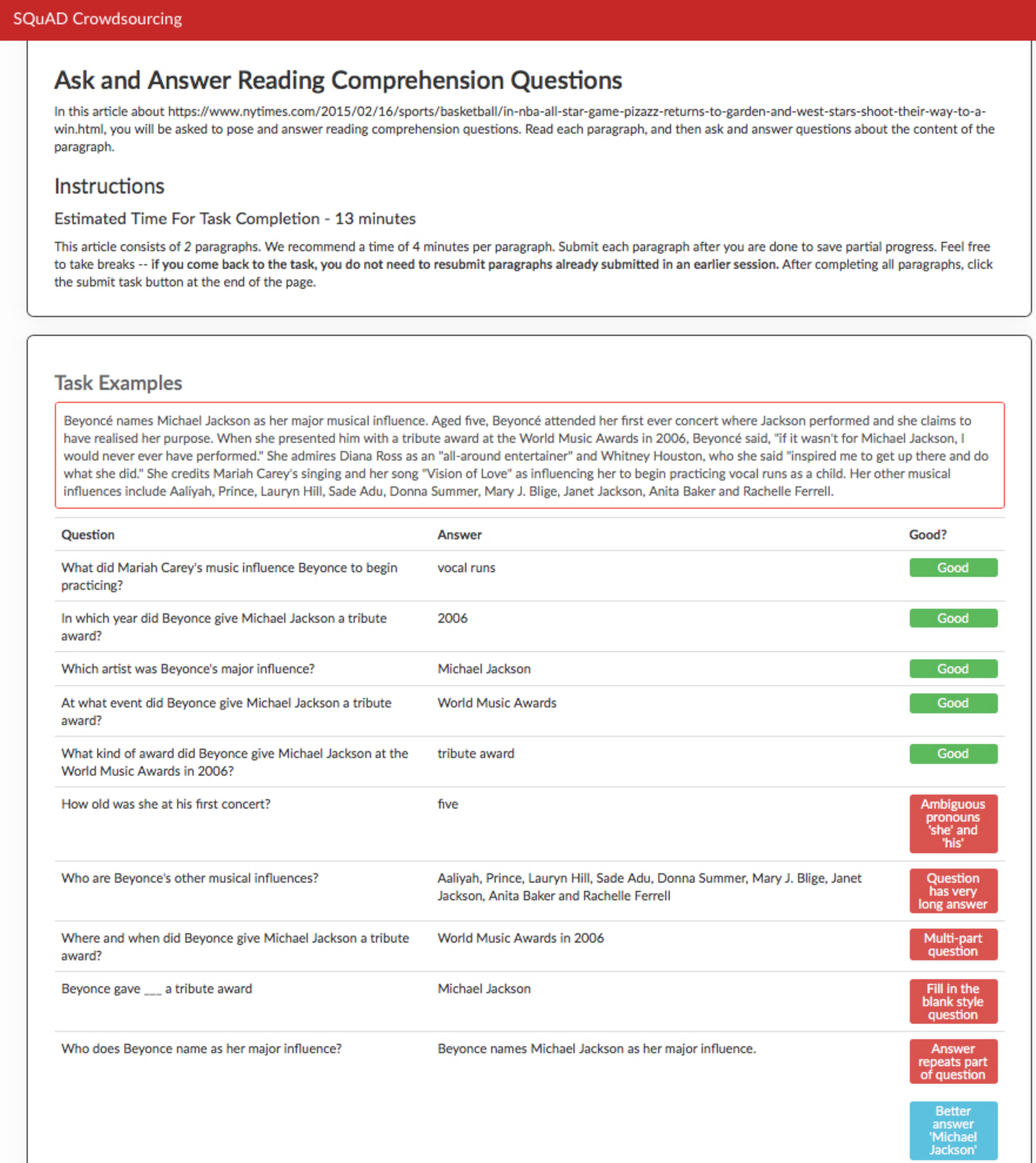}
    \caption{Ask task directions.}
        \label{fig:ask_directions}
\end{figure*}

\begin{figure*}[ht!]
    \centering
    \includegraphics[width=\linewidth]{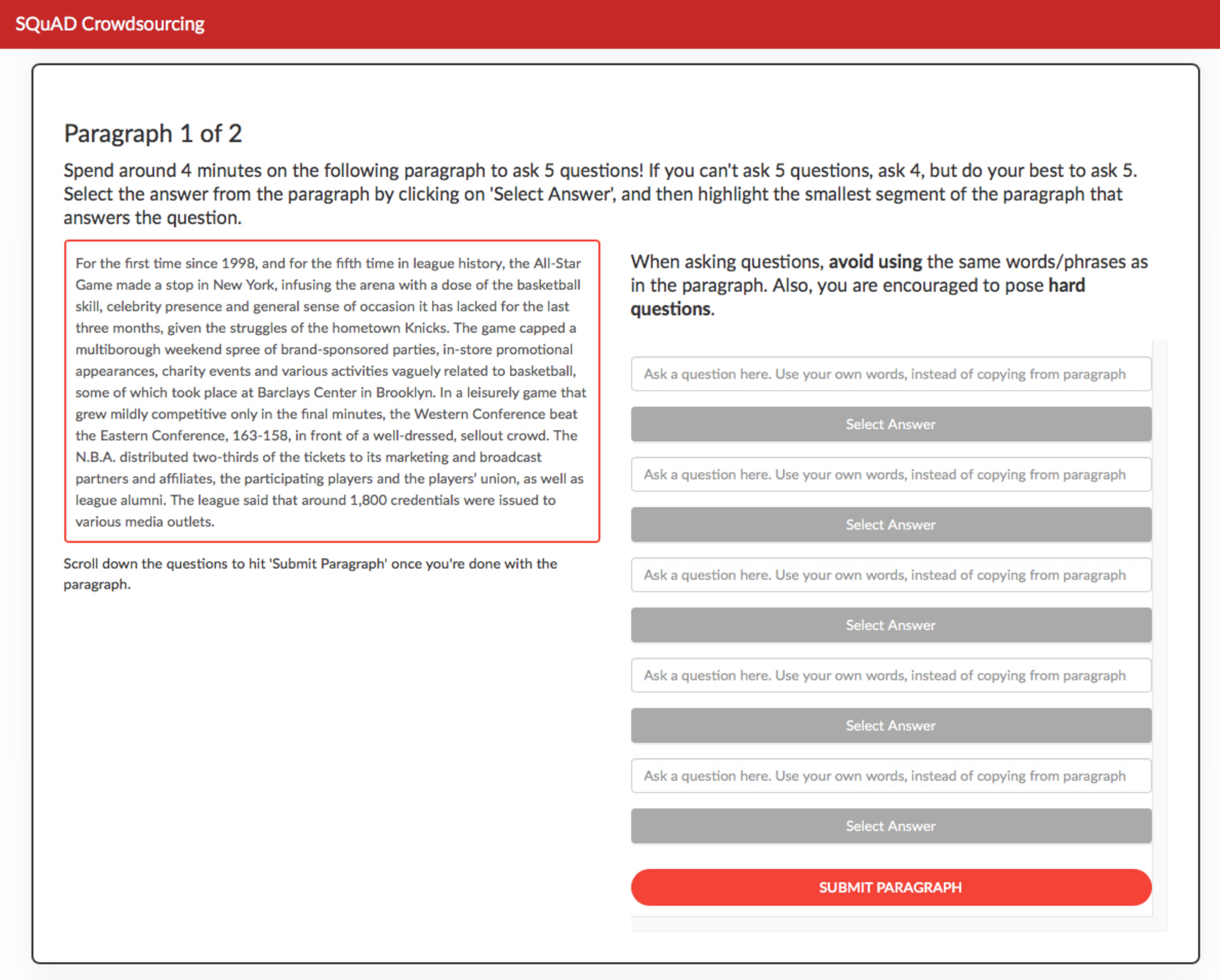}
    \caption{Ask task example.}
        \label{fig:ask_example}
\end{figure*}

\begin{figure*}[ht!]
    \centering
    \includegraphics[width=\linewidth]{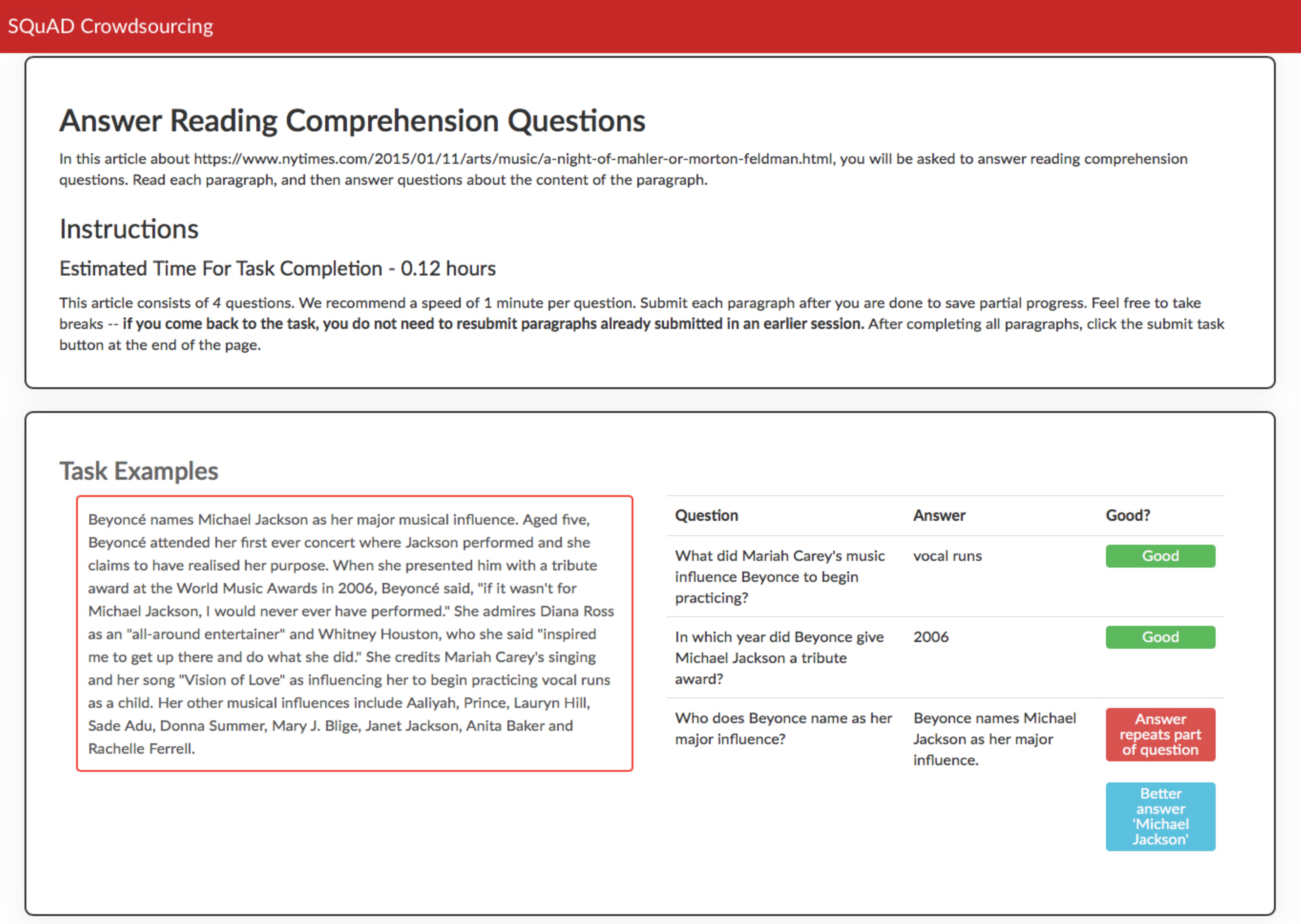}
    \caption{Answer task directions.}
        \label{fig:answer_directions}
\end{figure*}

\begin{figure*}[ht!]
    \centering
    \includegraphics[width=\linewidth]{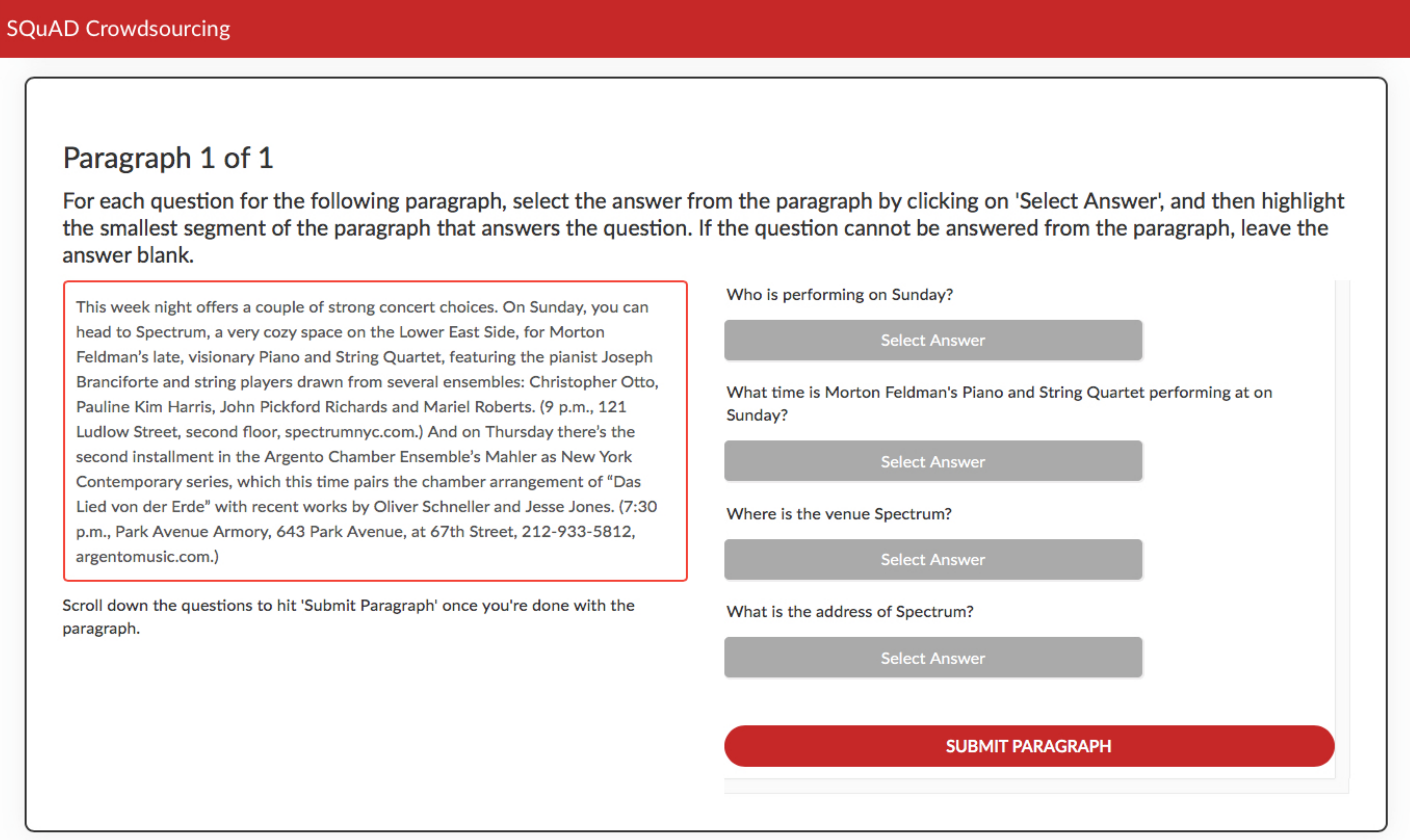}
    \caption{Answer task example.}
        \label{fig:answer_example}
\end{figure*}

\paragraph{UI Examples.}
The task directions and website UI are identical to the original SQuAD data
collection setup with the sole exception that the original tasks had workers ask
and answer questions for all of the paragraphs for each article, whereas our
tasks limit each worker to at most 5 paragraphs.
Figures~\ref{fig:ask_directions}~and~\ref{fig:ask_example} show the directions
and an example HIT for the Ask task, whereby workers pose questions for the
article.  Figures~\ref{fig:answer_directions}~and~\ref{fig:answer_example} show
the directions and an example HIT for the Answer task, whereby workers answer
questions posed during the Ask task.


\FloatBarrier

\section{Complete Model Testbed and Results Tables}
\label{app:full_results}
In this section, we detail the complete model testbed and provide evaluation
results for each model on each of our four distribution shift datasets, as well
as the adversarial distributions discussed in Section~\ref{app:adversarial}.

\subsection{Models Evaluated}
\label{app:model_list}
We evaluated a representative subset of over 100 models submitted to the \squad
leaderboard since 2016.  All of the models were submitted to the CodaLab
platform, and thus we evaluate every model in the exact same configuration
(weights, hyperparameters, command-line arguments, execution environment, etc.)
as the original submission. Below, we list all of the models we evaluated with
references, where available, and links to the Codalab submission bundle. The
models are listed in sorted order based on their \squad v1.1 Test F1 score to
allow easy reference to the subsequent tables.

{
\footnotesize
\begin{enumerate}
\item
XLNet (Single)~\citep{yang2019xlnet}\\
\url{https://worksheets.codalab.org/bundles/0x74ebcd1a59044db49472900ae9936cf3}
\item
XLNet-123 (Single)\\
\url{https://worksheets.codalab.org/bundles/0x519d3e06a3544b0e85b7477ea512ec01}
\item
XLNet-123++ (Single)\\
\url{https://worksheets.codalab.org/bundles/0x8a03e7cddcea47fa9395ca96870b62fd}
\item
SpanBERT (Single)~\citep{joshi2020spanbert}\\
\url{https://worksheets.codalab.org/bundles/0xe7315e3e35c64097af5351bb2dbdf9a5}
\item
BERT + WWM + MT (Single)\\
\url{https://worksheets.codalab.org/bundles/0x3975475041324f8c8b14626c932d09f4}
\item
Tuned BERT-1seq Large Cased (Single)~\citep{joshi2020spanbert}\\
\url{https://worksheets.codalab.org/bundles/0xa62618d05255460a83adfe1bfd1784f7}
\item
InfoWord Large (Single)~\citep{kong2019mutual}\\
\url{https://worksheets.codalab.org/bundles/0x4a19b4d7c2fb40ef913bd97f611e66bd}
\item
BERT-Large Baseline (single model)\\
\url{https://worksheets.codalab.org/bundles/0xcd68d4f224b0425ab2b8b34ffb140a75}
\item
BERT + MT (Single)\\
\url{https://worksheets.codalab.org/bundles/0x8e20cbb02fa64883afdb4f8e50357858}
\item
Tuned BERT Large Cased (Single)~\citep{devlin2018bert,joshi2020spanbert}\\
\url{https://worksheets.codalab.org/bundles/0x766e1c3149bd424fb154e31ee530845a}
\item
DPN (Single)\\
\url{https://worksheets.codalab.org/bundles/0xd362627c900146178b5c190161bf61cf}
\item
ST\_bl (Single)\\
\url{https://worksheets.codalab.org/bundles/0x79ca106fd7b5402abd3815636368ce2c}
\item
BERT uncased (Single)\\
\url{https://worksheets.codalab.org/bundles/0x1bbff660e00c4445a3dc11277039edc3}
\item
EL-BERT (Single)\\
\url{https://worksheets.codalab.org/bundles/0x2d65a49640394cba8632f765b237a41f}
\item
BISAN (single model)\\
\url{https://worksheets.codalab.org/bundles/0xfd43e046161f4ba89716d5d48b25ca2f}
\item
BERT + Sparse-Transformer (Single)\\
\url{https://worksheets.codalab.org/bundles/0xb1a4af82c1364cc1a41aef78543d0f52}
\item
InfoWord Base (Single)~\citep{kong2019mutual}\\
\url{https://worksheets.codalab.org/bundles/0xd2067806f74c4da79e81b73eca08bcba}
\item
InfoWord-Base (single model)\\
\url{https://worksheets.codalab.org/bundles/0xa41e1de495f84786a1c84d6f6036af0d}
\item
InfoWord BERT Large Baseline (Single)~\citep{devlin2018bert,kong2019mutual}\\
\url{https://worksheets.codalab.org/bundles/0x86fb7e7680b6488daa585dcd11e41a36}
\item
Original BERT Large Cased (Single)~\citep{devlin2018bert,joshi2020spanbert}\\
\url{https://worksheets.codalab.org/bundles/0x6603ef1196fd409d81948e3af7b44e58}
\item
Commonsense Governed BERT-123 (Single; May 8th)\\
\url{https://worksheets.codalab.org/bundles/0x8eecf515978a4fd382e077efecbf90e1}
\item
InfoWord BERT Base Baseline (Single)~\citep{devlin2018bert,kong2019mutual}\\
\url{https://worksheets.codalab.org/bundles/0xb6d9adf28e4241e181602540aeafa5a0}
\item
Commonsense Governed BERT-123 (Single; April 21st)\\
\url{https://worksheets.codalab.org/bundles/0x008044bbd7f74a7a81b51cbcfdf5a654}
\item
MARS (Ensemble; June 20th)\\
\url{https://worksheets.codalab.org/bundles/0xb320588e9f424639b54f1f40de9b0cf9}
\item
MARS (Single; September 1st)\\
\url{https://worksheets.codalab.org/bundles/0xfaf7cb0df0af4bf5a4050a53b81be174}
\item
MARS (Single; June 21st)\\
\url{https://worksheets.codalab.org/bundles/0xfc0c5b744d2a4c6b9f709c98bc2cf4e9}
\item
MMIPN (Single)\\
\url{https://worksheets.codalab.org/bundles/0xc2c7813ec5e241e2a0c43da45c7ecc91}
\item
MARS (Single; May 9th)\\
\url{https://worksheets.codalab.org/bundles/0x6d7c8a0f92374218ab4d419b397d67eb}
\item
Reinforced Mnemonic Reader (Ensemble)~\citep{hu2018reinforced}\\
\url{https://worksheets.codalab.org/bundles/0x0a5ea1308bad49b2bcdd37250bdf844a}
\item
AttentionReader+ (Ensemble)\\
\url{https://worksheets.codalab.org/bundles/0x50985a93bf734c40b76b8cc915fe967b}
\item
Reinforced Mnemonic Reader + A2D (Single)\\
\url{https://worksheets.codalab.org/bundles/0x6ada3ab4807442a4944de1e8ec1f5681}
\item
Reinforced Mnemonic Reader + A2D + DA (Single)\\
\url{https://worksheets.codalab.org/bundles/0xeb52c2067dca498d852ca693eb9fd68a}
\item
BERT-Compound-DSS (Single)\\
\url{https://worksheets.codalab.org/bundles/0xd74488aac2e04d47983cbee5e7a8a106}
\item
BERT-Compound (Single)\\
\url{https://worksheets.codalab.org/bundles/0xc0dc1a25c03e4ba493ec28eef0e643b6}
\item
BiDAF + Self-Attention + ELMo (Ensemble)~\citep{peters2018deep}\\
\url{https://worksheets.codalab.org/bundles/0x35b427e3105a46498256e3ccd502e442}
\item
AVIQA+ (Ensemble)\\
\url{https://worksheets.codalab.org/bundles/0x0109d51630ac45599a85523d4690afd1}
\item
EAZI (Ensemble)\\
\url{https://worksheets.codalab.org/bundles/0x55c1434feb8d48dfb990756ee1ce86d8}
\item
EAZI+ (Ensemble)\\
\url{https://worksheets.codalab.org/bundles/0x0b44f79d1e8042dd94943a35a057d7ea}
\item
MEMEN+ (Ensemble)\\
\url{https://worksheets.codalab.org/bundles/0x065d328704784db7b093f3e750ff1b46}
\item
DNET (Ensemble)\\
\url{https://worksheets.codalab.org/bundles/0x5b80aaba5fde4f65823746bb9b8a8fdc}
\item
BERT-Independent (Single)\\
\url{https://worksheets.codalab.org/bundles/0x82178b8ab098491aabec5b3a1ed18994}
\item
Reinforced Mnemonic Reader (Single)~\citep{hu2018reinforced}\\
\url{https://worksheets.codalab.org/bundles/0x78c31b2a1b9846a4b9de7dd71124656b}
\item
FusionNet (Ensemble)~\citep{huang2017fusionnet}\\
\url{https://worksheets.codalab.org/bundles/0xd4ff6ed2458e4df099ea677a20115128}
\item
MDReader (Single)\\
\url{https://worksheets.codalab.org/bundles/0xed0bb85059b04ce79db37982b1381801}
\item
BiDAF + Self Attention + ELMo (single model)\\
\url{https://worksheets.codalab.org/bundles/0x11f631b3e7cb4a0f8acbd60491f729b6}
\item
BiDAF + Self-Attention + ELMo (Single)~\citep{peters2018deep}\\
\url{https://worksheets.codalab.org/bundles/0x5ab1fa7d11f04c5991d5011471ebdc4c}
\item
MDReader0 (Single)\\
\url{https://worksheets.codalab.org/bundles/0x17bde05ef4b4483a9acf9e1ef8cc9326}
\item
BiDAF++ + pair2vec (Single)~\citep{joshi2019pair2vec}\\
\url{https://worksheets.codalab.org/bundles/0x1720fa746b0243e19692820fd930b14e}
\item
Conductor-net (Ensemble)~\citep{liu2017phase}\\
\url{https://worksheets.codalab.org/bundles/0x21d981f8667141b5bf6871714e3d5fd2}
\item
MEMEN+ (Single)\\
\url{https://worksheets.codalab.org/bundles/0xf4709036e11843f88f870f4e7dea50a0}
\item
AVIQA v2 (Ensemble)\\
\url{https://worksheets.codalab.org/bundles/0x796847815444478c842f63a97cef93a0}
\item
MEMEN (Single; model submitted after paper)~\citep{pan2017memen}\\
\url{https://worksheets.codalab.org/bundles/0x55fcc3f13d664944969bf05c59f402a4}
\item
Interactive AoA Reader (Ensemble)\\
\url{https://worksheets.codalab.org/bundles/0x00599dfa3921413cab3a75a70722234d}
\item
EAZI (single model)\\
\url{https://worksheets.codalab.org/bundles/0xad2056e99a0a484f8b8e4bcc2b1b0c14}
\item
AttentionReader+ (Single)\\
\url{https://worksheets.codalab.org/bundles/0x334adb7624674e90aff7be232fb52005}
\item
DNET (Single)\\
\url{https://worksheets.codalab.org/bundles/0x5eb36fb24feb4911888760f8554f90ac}
\item
BiDAF++ (Single)~\citep{joshi2019pair2vec}\\
\url{https://worksheets.codalab.org/bundles/0xb9a6b77b0163453c8fb942bafa1e2cfe}
\item
MARS (Single; January 23rd)\\
\url{https://worksheets.codalab.org/bundles/0x92ce58765d194debbadc1a165399a454}
\item
FRC (Single)\\
\url{https://worksheets.codalab.org/bundles/0x346b188552ed4d1cb6c1bccaa6d243eb}
\item
Jenga (Ensemble)\\
\url{https://worksheets.codalab.org/bundles/0xbc23efc53a1f4735bad72aa01546ace1}
\item
RaSoR + TR + LM (Single)~\citep{salant2018contextualized}\\
\url{https://worksheets.codalab.org/bundles/0xec9321a11b0f44e19ca8d325dcda75eb}
\item
{gqa} (single model)\\
\url{https://worksheets.codalab.org/bundles/0xc8548cd7df0547dd9003a02e5505dd77}
\item
FusionNet (Single)~\citep{huang2017fusionnet}\\
\url{https://worksheets.codalab.org/bundles/0xbe9fefbe5b544675aafee4e83ccbe1e1}
\item
Smarnet (Ensemble)~\citep{chen2017smarnet}\\
\url{https://worksheets.codalab.org/bundles/0x622060479ede4552bf490c942598ac3c}
\item
AVIQA v2 (Single)\\
\url{https://worksheets.codalab.org/bundles/0x58ce6e7730b241dea20597b0a0e51b7e}
\item
DCN+ (Single)~\citep{xiong2017dcn+}\\
\url{https://worksheets.codalab.org/bundles/0xd38944b81f484cf6a40955778204a0cf}
\item
Jenga (single model)\\
\url{https://worksheets.codalab.org/bundles/0x38bce62d659e43d19f56fc2ba34c3c4d}
\item
MixedModel (Ensemble)\\
\url{https://worksheets.codalab.org/bundles/0x761449f9e327450e85938688a002bc72}
\item
Two-Attention + Self-Attention (Ensemble)\\
\url{https://worksheets.codalab.org/bundles/0xf9087be2e1a34b96809b71e8ccaf9c56}
\item
MEMEN (Ensemble; original model in paper)~\citep{pan2017memen}\\
\url{https://worksheets.codalab.org/bundles/0x5596d3b1dceb414eab5653c5ec8f1607}
\item
ReasoNet (Ensemble)~\citep{shen2017reasonet}\\
\url{https://worksheets.codalab.org/bundles/0xe117260a328f484590e34b91839ce9ad}
\item
eeAttNet (Single)\\
\url{https://worksheets.codalab.org/bundles/0x48a65548231d47a1aed7f5554f724064}
\item
Mnemonic Reader (Ensemble)~\citep{hu2018reinforced}\\
\url{https://worksheets.codalab.org/bundles/0xa860db3ea8854156b68da2e3a9a2f962}
\item
Conductor-net (Single)~\citep{liu2017phase}\\
\url{https://worksheets.codalab.org/bundles/0x6fce3642dc574820949b0ae40bbac564}
\item
Interactive AoA Reader (Single)\\
\url{https://worksheets.codalab.org/bundles/0x6541c8fd5acb44cf85572d6827c22f44}
\item
Jenga (Single)\\
\url{https://worksheets.codalab.org/bundles/0x4b25320ab45d459fb4274c15ed925322}
\item
SSAE (Ensemble)\\
\url{https://worksheets.codalab.org/bundles/0x34a9c6dd5f3145ce9130ddba8a951254}
\item
jNet (Ensemble)~\citep{zhang2017exploring}\\
\url{https://worksheets.codalab.org/bundles/0x9ba8c5bbe77c4fd399d670ca11e42695}
\item
BiDAF + Self-Attention (Single)~\citep{clark2018simple}\\
\url{https://worksheets.codalab.org/bundles/0xe0b60a2436ef407cbf5fa0641c5350ba}
\item
Two-Attention + Self-Attention (Single)\\
\url{https://worksheets.codalab.org/bundles/0xfcb73b26ac0049478c0b4ae4f09cb3c9}
\item
AVIQA (Single)\\
\url{https://worksheets.codalab.org/bundles/0x513d75fb3d554dd6bc11dafb7ef1f5c3}
\item
Attention + Self-Attention (Single)\\
\url{https://worksheets.codalab.org/bundles/0xbd549e52d11b42b39bd3d2fc0bbbe1da}
\item
Smarnet (Single)~\citep{chen2017smarnet}\\
\url{https://worksheets.codalab.org/bundles/0x733cef4d589743b8bc95a6108206c8a0}
\item
Mnemonic Reader (Single)~\citep{hu2018reinforced}\\
\url{https://worksheets.codalab.org/bundles/0x28ff5339d7164a2ea95db1a4a3a2a750}
\item
MAMCN (Single)~\citep{yu2018multi}\\
\url{https://worksheets.codalab.org/bundles/0x3d6ebcc7d54d44798d477e94fc840830}
\item
M-NET (Single)\\
\url{https://worksheets.codalab.org/bundles/0x978c1865473f4a34bf23c14b152ec4e1}
\item
jNet (Single)~\citep{zhang2017exploring}\\
\url{https://worksheets.codalab.org/bundles/0x8c62efeae93743018965441fe6e7ced0}
\item
Ruminating Reader (Single)~\citep{gong2018ruminating}\\
\url{https://worksheets.codalab.org/bundles/0x5abfb433377c45f3b6e3d26c3f6cd050}
\item
ReasoNet (Single)~\citep{shen2017reasonet}\\
\url{https://worksheets.codalab.org/bundles/0x2356880cbc5347069d99a8cf38815dbc}
\item
RaSoR (Single)~\citep{lee2016learning}\\
\url{https://worksheets.codalab.org/bundles/0x9dba642677a4489eb8fc78969601c893}
\item
SimpleBaseline (Single)\\
\url{https://worksheets.codalab.org/bundles/0xd78f5da9c45d4fa5bde361f9370b8a40}
\item
PQMN (Single)\\
\url{https://worksheets.codalab.org/bundles/0x0f29cad4f3e94dcfb4560e4347d946d5}
\item
AllenNLP BiDAF (Single)~\citep{seo2016bidirectional,gardner2018allennlp}\\
\url{https://worksheets.codalab.org/bundles/0x8704f9226d884b5687fba7f73a462195}
\item
Match-LSTM w/ Ans-Ptr Boundary (Ensemble)~\citep{wang2016machine}\\
\url{https://worksheets.codalab.org/bundles/0x0bbda0093b294c1191a9dda91c0aa9b0}
\item
Iterative Co-Attention Network (Single)\\
\url{https://worksheets.codalab.org/bundles/0x801a86cd3dbd44ae930c7134b7ababe5}
\item
BiDAF-Compound-DSS (Single)\\
\url{https://worksheets.codalab.org/bundles/0xc46b10050145494aa93708faa40b4013}
\item
BiDAF-Independent-DSS (Single)\\
\url{https://worksheets.codalab.org/bundles/0x2254478ccad84effbd92de915ff063be}
\item
BiDAF-Independent (Single)\\
\url{https://worksheets.codalab.org/bundles/0x3d6cd49604b8466ca952fda73bfb2527}
\item
BiDAF-Compound (Single)\\
\url{https://worksheets.codalab.org/bundles/0x450cd98f9ab548049b8e28c9f225910e}
\item
Match-LSTM w/ Bi-Ans-Ptr Boundary (Single)~\citep{wang2016machine}\\
\url{https://worksheets.codalab.org/bundles/0x5f678f88703f4eb0b320793ed998dc20}
\item
OTF Dict + Spelling (Single)~\citep{bahdanau2017learning}\\
\url{https://worksheets.codalab.org/bundles/0xd33f2fbd7eca4819b2c2b45371abcdf4}
\item
OTF Spelling (Single)~\citep{bahdanau2017learning}\\
\url{https://worksheets.codalab.org/bundles/0x5ce7b655beb0454da5240c17f36bce6c}
\item
OTF Spelling + Lemma (Single)~\citep{bahdanau2017learning}\\
\url{https://worksheets.codalab.org/bundles/0x308cfd9f735d4965835ec496610ea91d}
\item
RQA+IDR (single model)\\
\url{https://worksheets.codalab.org/bundles/0x54e292cee87d4b1488b9cf0df15aeeec}
\item
Dynamic Chunk Reader (Single)~\citep{yu2016end}\\
\url{https://worksheets.codalab.org/bundles/0x345be18cbe4541de841de3ac79d5b441}
\item
UQA (single model)\\
\url{https://worksheets.codalab.org/bundles/0x64206b3164ea47e7a3d8a2df833c8f9b}
\item
UnsupervisedQA V1\\
\url{https://worksheets.codalab.org/bundles/0xe1c53a62c8644e9b9d9fdfd18feb6a85}
\end{enumerate}

}

We also evaluated a subset of five models from the Machine Reading for Question
Answering (MRQA) Shared Task~\citep{fisch2019mrqa} on our new test sets. As in
our primary experiments, all of the models were submitted to the CodaLab
platform, and  we evaluated every model in the exact same configuration as the
original submission. Below, we list all of the models we evaluated with
references and links to the submission bundle.
\begin{enumerate}
\item 
{\tt Delphi}~\citep{longpre2019exploration}\\
\url{https://worksheets.codalab.org/bundles/0x9a53e9c50f1244699c4a24aee483bd4c}
\item
{\tt HierAtt}~\citep{osama2019question} \\
\url{https://worksheets.codalab.org/bundles/0x8d851db3255b485c97646c5c0ba812a2}
\item
{\tt Bert-Large+Adv Train}~\citep{lee2019domain}\\
\url{https://worksheets.codalab.org/bundles/0xa113983bc3fc42ff89bf3838a6177a0c}
\item
{\tt BERT-cased-whole-word}\\
\url{https://worksheets.codalab.org/bundles/0x456676760aae452cb44ade00bb515b64}
\item
{\tt BERT-Multi-Finetune}\\
\url{https://worksheets.codalab.org/bundles/0x5716df3b477a452a997bcebb9e179c89}
\end{enumerate}
The remaining models submitted to the competition were either not publicly
accessible or otherwise unable to run on Codalab.

\subsection{Full Results Tables}
\label{app:results_tables}
\paragraph{Main Results.}
In this section, we present the results for each \squad model and the
5 MRQA model listed in Appendix~\ref{app:model_list}, along with results for the
three student and postdoc authors of this paper, on each of our new test sets.
Tables~\ref{table:full_new_wiki_f1_table},~\ref{table:full_nyt_f1_table}
~\ref{table:full_reddit_f1_table}, and~\ref{table:full_amazon_f1_table} contain the
results for each our models and the three human annotators in terms of F1 score
for the New Wikipedia, New York Times, Reddit, and Amazon test sets,
respectively. Tables~\ref{table:full_new_wiki_em_table},
~\ref{table:full_nyt_em_table},~\ref{table:full_reddit_em_table},
and~\ref{table:full_amazon_em_table} contain the same data for exact match
scores. For a particular dataset, some models are not listed if we were unable
to evaluate the model on the dataset in Codalab.

\begin{center}
    \footnotesize
    \rowcolors{2}{gray!15}{white}
\begin{longtable}{llccccc}
    \label{table:full_new_wiki_f1_table}\\
    \caption{
        Comparison of model F1 scores on the original SQuAD test set and our
        new Wikipedia test set. Rank refers to the relative ordering using the
        original \squad v1.1 F1 scores, new rank refers to the ordering using
        the new test set scores, and $\Delta$ rank is the relative difference in
        ranking.  The confidence intervals are 95\% Student's t-intervals. Unless
        noted, all models are {\tt single models} and only the first 30
        characters of each name is shown.
    } \\
    \appendixtable{New Wikipedia}{F1}
    	- & Human-0 & 94.9  \textcolor{gray!90}{[93.8, 96.0]} & 92.5  \textcolor{gray!90}{[91.1, 93.8]} & 2.4 & - & - \\
	- & Human-1 & 94.9  \textcolor{gray!90}{[93.8, 96.0]} & 92.4  \textcolor{gray!90}{[91.0, 93.8]} & 2.5 & - & - \\
	- & Human-2 & 95.6  \textcolor{gray!90}{[94.5, 96.6]} & 92.3  \textcolor{gray!90}{[90.8, 93.8]} & 3.2 & - & - \\
	1 & XLNet (single model) & 95.1 & 92.3  \textcolor{gray!90}{[91.9, 92.8]} & 2.7 & 1 & 0 \\
	2 & XLNET-123 (single model) & 94.9 & 92.2  \textcolor{gray!90}{[91.7, 92.7]} & 2.7 & 5 & -3 \\
	3 & XLNET-123++ (single model) & 94.9 & 92.3  \textcolor{gray!90}{[91.8, 92.7]} & 2.6 & 3 & 0 \\
\end{longtable}
\end{center}

\begin{center}
    \footnotesize
    \rowcolors{2}{gray!15}{white}
    \begin{longtable}{llccccc}
    \caption{
        Comparison of model F1 scores on the original SQuAD test set and our
        New York Times test set . Rank refers to the relative ordering using the
        original \squad v1.1 F1 scores, new rank refers to the ordering using
        the new test set scores, and $\Delta$ rank is the relative difference in
        ranking.  The confidence intervals are 95\% Student's t-intervals. Unless
        noted, all models are {\tt single models} and only the first 30
        characters of each name is shown.
    }\\
    \appendixtable{NYT}{F1}
    	- & Human-0 & 94.9  \textcolor{gray!90}{[93.8, 96.0]} & 95.0  \textcolor{gray!90}{[93.9, 96.1]} & -0.1 & - & - \\
	- & Human-1 & 94.9  \textcolor{gray!90}{[93.8, 96.0]} & 96.3  \textcolor{gray!90}{[95.4, 97.1]} & -1.4 & - & - \\
	- & Human-2 & 95.6  \textcolor{gray!90}{[94.5, 96.6]} & 93.7  \textcolor{gray!90}{[92.4, 95.0]} & 1.9 & - & - \\
	1 & XLNet (single model) & 95.1 & 84.4  \textcolor{gray!90}{[84.0, 84.9]} & 10.7 & 34 & -33 \\
	2 & XLNET-123 (single model) & 94.9 & 92.8  \textcolor{gray!90}{[92.3, 93.2]} & 2.2 & 3 & -1 \\
	3 & XLNET-123++ (single model) & 94.9 & 92.9  \textcolor{gray!90}{[92.5, 93.3]} & 2.0 & 2 & 1 \\
	4 & Delphi & 94.7 & 93.4  \textcolor{gray!90}{[93.0, 93.8]} & 1.3 & 1 & 3 \\
\end{longtable}
\end{center}

\begin{center}
    \footnotesize
    \rowcolors{2}{gray!15}{white}
    \begin{longtable}{llccccc}
    \caption{
        Comparison of model F1 scores on the original SQuAD test set and our
        Reddit test set. Rank refers to the relative ordering using the
        original \squad v1.1 F1 scores, new rank refers to the ordering using
        the new test set scores, and $\Delta$ rank is the relative difference in
        ranking.  The confidence intervals are 95\% Student's t-intervals. Unless
        noted, all models are {\tt single models} and only the first 30
        characters of each name is shown.
    }\\
    \appendixtable{Reddit}{F1}
    	- & Human-0 & 94.9  \textcolor{gray!90}{[93.8, 96.0]} & 92.4  \textcolor{gray!90}{[91.1, 93.7]} & 2.5 & - & - \\
	- & Human-1 & 94.9  \textcolor{gray!90}{[93.8, 96.0]} & 92.6  \textcolor{gray!90}{[91.3, 93.9]} & 2.3 & - & - \\
	- & Human-2 & 95.6  \textcolor{gray!90}{[94.5, 96.6]} & 91.7  \textcolor{gray!90}{[90.2, 93.2]} & 3.8 & - & - \\
	1 & XLNet (single model) & 95.1 & 79.0  \textcolor{gray!90}{[78.5, 79.6]} & 16.0 & 21 & -20 \\
	2 & XLNET-123 (single model) & 94.9 & 84.9  \textcolor{gray!90}{[84.2, 85.5]} & 10.1 & 3 & -1 \\
	3 & XLNET-123++ (single model) & 94.9 & 84.8  \textcolor{gray!90}{[84.2, 85.4]} & 10.1 & 4 & -1 \\
	4 & Delphi & 94.7 & 88.0  \textcolor{gray!90}{[87.5, 88.6]} & 6.7 & 1 & 3 \\
\end{longtable}
\end{center}

\begin{center}
    \footnotesize
    \rowcolors{2}{gray!15}{white}
    \begin{longtable}{llccccc}
    \caption{
        Comparison of model F1 scores on the original SQuAD test set and our
        Amazon test set. Rank refers to the relative ordering using the
        original \squad v1.1 F1 scores, new rank refers to the ordering using
        the new test set scores, and $\Delta$ rank is the relative difference in
        ranking.  The confidence intervals are 95\% Student's t-intervals. Unless
        noted, all models are {\tt single models} and only the first 30
        characters of each name is shown.
    }\\
    \appendixtable{Amazon}{F1}
    	- & Human-0 & 94.9  \textcolor{gray!90}{[93.8, 96.0]} & 92.6  \textcolor{gray!90}{[91.3, 93.9]} & 2.3 & - & - \\
	- & Human-1 & 94.9  \textcolor{gray!90}{[93.8, 96.0]} & 92.4  \textcolor{gray!90}{[91.1, 93.7]} & 2.5 & - & - \\
	- & Human-2 & 95.6  \textcolor{gray!90}{[94.5, 96.6]} & 91.2  \textcolor{gray!90}{[89.6, 92.8]} & 4.4 & - & - \\
	1 & XLNet (single model) & 95.1 & 81.7  \textcolor{gray!90}{[81.1, 82.2]} & 13.4 & 7 & -6 \\
	2 & XLNET-123 (single model) & 94.9 & 85.7  \textcolor{gray!90}{[85.1, 86.3]} & 9.2 & 3 & -1 \\
	3 & XLNET-123++ (single model) & 94.9 & 87.2  \textcolor{gray!90}{[86.7, 87.7]} & 7.7 & 2 & 1 \\
	4 & Delphi & 94.7 & 87.7  \textcolor{gray!90}{[87.2, 88.3]} & 6.9 & 1 & 3 \\
\end{longtable}
\end{center}

\begin{center}
    \footnotesize
    \rowcolors{2}{gray!15}{white}
    \begin{longtable}{llccccc}
    \caption{
        Comparison of model EM scores on the original SQuAD test set and our
        new Wikipedia test set. Rank refers to the relative ordering using the
        original \squad v1.1 EM scores, new rank refers to the ordering using
        the new test set scores, and $\Delta$ rank is the relative difference in
        ranking.  The confidence intervals are 95\% Clopper-Pearson intervals.
        Unless noted, all models are {\tt single models} and only the first 30
        characters of each name is shown.
    }\\
    \appendixtable{New Wiki}{EM}
    	- & Human-0 & 89.1  \textcolor{gray!90}{[87.1, 91.0]} & 82.6  \textcolor{gray!90}{[80.0, 85.0]} & 6.6 & - & - \\
	- & Human-1 & 88.7  \textcolor{gray!90}{[86.6, 90.6]} & 83.2  \textcolor{gray!90}{[80.6, 85.6]} & 5.5 & - & - \\
	- & Human-2 & 90.5  \textcolor{gray!90}{[88.5, 92.2]} & 85.4  \textcolor{gray!90}{[82.9, 87.6]} & 5.1 & - & - \\
	1 & XLNet (single model) & 89.9 & 84.2  \textcolor{gray!90}{[83.3, 85.0]} & 5.7 & 1 & 0 \\
	2 & XLNET-123++ (single model) & 89.9 & 83.4  \textcolor{gray!90}{[82.6, 84.2]} & 6.4 & 5 & -3 \\
	3 & XLNET-123 (single model) & 89.6 & 83.5  \textcolor{gray!90}{[82.7, 84.3]} & 6.1 & 3 & 0 \\
	4 & Delphi & 89.6 & 84.0  \textcolor{gray!90}{[83.1, 84.8]} & 5.6 & 2 & 2 \\
\end{longtable}
\end{center}

\begin{center}
    \footnotesize
    \rowcolors{2}{gray!15}{white}
    \begin{longtable}{llccccc}
    \caption{
        Comparison of model EM scores on the original SQuAD test set and our
        New York Times test set. Rank refers to the relative ordering using the
        original \squad v1.1 EM scores, new rank refers to the ordering using
        the new test set scores, and $\Delta$ rank is the relative difference in
        ranking.  The confidence intervals are 95\% Clopper-Pearson intervals.
        Unless noted, all models are {\tt single models} and only the first 30
        characters of each name is shown.
    } \\
    \appendixtable{NYT}{EM}
    	- & Human-0 & 89.1  \textcolor{gray!90}{[87.1, 91.0]} & 86.0  \textcolor{gray!90}{[83.6, 88.1]} & 3.2 & - & - \\
	- & Human-1 & 88.7  \textcolor{gray!90}{[86.6, 90.6]} & 88.5  \textcolor{gray!90}{[86.3, 90.5]} & 0.2 & - & - \\
	- & Human-2 & 90.5  \textcolor{gray!90}{[88.5, 92.2]} & 85.8  \textcolor{gray!90}{[83.4, 87.9]} & 4.7 & - & - \\
	1 & XLNet (single model) & 89.9 & 50.9  \textcolor{gray!90}{[50.0, 51.9]} & 38.9 & 102 & -101 \\
	2 & XLNET-123++ (single model) & 89.9 & 85.9  \textcolor{gray!90}{[85.2, 86.6]} & 3.9 & 3 & -1 \\
	3 & XLNET-123 (single model) & 89.6 & 86.0  \textcolor{gray!90}{[85.3, 86.7]} & 3.7 & 2 & 1 \\
	4 & Delphi & 89.6 & 86.9  \textcolor{gray!90}{[86.3, 87.6]} & 2.7 & 1 & 3 \\
\end{longtable}
\end{center}

\begin{center}
    \footnotesize
    \rowcolors{2}{gray!15}{white}
    \begin{longtable}{llccccc}
    \caption{
        Comparison of model EM scores on the original SQuAD test set and our
        Reddit test set. Rank refers to the relative ordering using the
        original \squad v1.1 EM scores, new rank refers to the ordering using
        the new test set scores, and $\Delta$ rank is the relative difference in
        ranking.  The confidence intervals are 95\% Clopper-Pearson intervals.
        Unless noted, all models are {\tt single models} and only the first 30
        characters of each name is shown.
    }\\
    \appendixtable{Reddit}{EM}
    	- & Human-0 & 89.1  \textcolor{gray!90}{[87.1, 91.0]} & 80.1  \textcolor{gray!90}{[77.2, 82.7]} & 9.1 & - & - \\
	- & Human-1 & 88.7  \textcolor{gray!90}{[86.6, 90.6]} & 80.7  \textcolor{gray!90}{[77.8, 83.3]} & 8.0 & - & - \\
	- & Human-2 & 90.5  \textcolor{gray!90}{[88.5, 92.2]} & 81.0  \textcolor{gray!90}{[78.2, 83.6]} & 9.4 & - & - \\
	1 & XLNet (single model) & 89.9 & 43.1  \textcolor{gray!90}{[42.1, 44.1]} & 46.8 & 89 & -88 \\
	2 & XLNET-123++ (single model) & 89.9 & 70.8  \textcolor{gray!90}{[69.9, 71.7]} & 19.0 & 5 & -3 \\
	3 & XLNET-123 (single model) & 89.6 & 73.8  \textcolor{gray!90}{[72.9, 74.7]} & 15.8 & 2 & 1 \\
	4 & Delphi & 89.6 & 77.9  \textcolor{gray!90}{[77.1, 78.8]} & 11.7 & 1 & 3 \\
\end{longtable}
\end{center}

\begin{center}
    \footnotesize
    \rowcolors{2}{gray!15}{white}
    \begin{longtable}{llccccc}
    \caption{
        Comparison of model EM scores on the original SQuAD test set and our
        Amazon test set. Rank refers to the relative ordering using the
        original \squad v1.1 EM scores, new rank refers to the ordering using
        the new test set scores, and $\Delta$ rank is the relative difference in
        ranking.  The confidence intervals are 95\% Clopper-Pearson intervals.
        Unless noted, all models are {\tt single models} and only the first 30
        characters of each name is shown.
    }\\
    \appendixtable{Amazon}{EM}
    	- & Human-0 & 89.1  \textcolor{gray!90}{[87.1, 91.0]} & 79.9  \textcolor{gray!90}{[77.1, 82.4]} & 9.3 & - & - \\
	- & Human-1 & 88.7  \textcolor{gray!90}{[86.6, 90.6]} & 81.1  \textcolor{gray!90}{[78.4, 83.6]} & 7.6 & - & - \\
	- & Human-2 & 90.5  \textcolor{gray!90}{[88.5, 92.2]} & 79.3  \textcolor{gray!90}{[76.5, 82.0]} & 11.2 & - & - \\
	1 & XLNet (single model) & 89.9 & 53.9  \textcolor{gray!90}{[52.9, 54.9]} & 36.0 & 53 & -52 \\
	2 & XLNET-123++ (single model) & 89.9 & 74.1  \textcolor{gray!90}{[73.2, 75.0]} & 15.7 & 2 & 0 \\
	3 & XLNET-123 (single model) & 89.6 & 72.2  \textcolor{gray!90}{[71.3, 73.1]} & 17.5 & 3 & 0 \\
	4 & Delphi & 89.6 & 75.7  \textcolor{gray!90}{[74.8, 76.5]} & 13.9 & 1 & 3 \\
\end{longtable}
\end{center}

\paragraph{Adversarial Results.}
The model evaluation data used to construct the adversarial plots from
Appendix~\ref{app:adversarial} is summarized in
Tables~\ref{table:full_addsent_f1_table}
and~\ref{table:full_addonesent_f1_table} for F1 scores and
Tables~\ref{table:full_addsent_em_table}
and~\ref{table:full_addonesent_em_table} for EM scores.

\begin{center}
    \footnotesize
    \rowcolors{2}{gray!15}{white}
    \begin{longtable}{llccccc}
    \caption{Model F1 scores for against the \emph{adversarial} distribution
    shifts {\tt AddSent}.}\\
    \appendixtable{AddSent}{F1}
    	1 & XLNet (single model) & 95.1 & 76.5 & 18.6 & 2 & -1 \\
	2 & XLNET-123 (single model) & 94.9 & 68.9 & 26.0 & 8 & -6 \\
	3 & XLNET-123++ (single model) & 94.9 & 77.2 & 17.7 & 1 & 2 \\
	4 & SpanBERT (single model) & 94.6 & 71.5 & 23.2 & 4 & 0 \\
	5 & BERT+WWM+MT & 94.4 & 73.9 & 20.5 & 3 & 2 \\
	6 & Tuned BERT-1seq Large Cased (s & 93.3 & 71.2 & 22.1 & 5 & 1 \\
	7 & InfoWord (large) & 93.1 & 64.4 & 28.7 & 12 & -5 \\
\end{longtable}
\end{center}

\begin{center}
    \footnotesize
    \rowcolors{2}{gray!15}{white}
    \begin{longtable}{llccccc}
    \caption{Model F1 scores for against the \emph{adversarial} distribution
    shifts {\tt AddOneSent}.}\\
    \appendixtable{AddOneSent}{F1}
    	1 & XLNet (single model) & 95.1 & 83.5 & 11.6 & 2 & -1 \\
	2 & XLNET-123 (single model) & 94.9 & 78.4 & 16.5 & 7 & -5 \\
	3 & XLNET-123++ (single model) & 94.9 & 84.5 & 10.4 & 1 & 2 \\
	4 & SpanBERT (single model) & 94.6 & 79.8 & 14.9 & 4 & 0 \\
	5 & BERT+WWM+MT & 94.4 & 80.8 & 13.6 & 3 & 2 \\
	6 & Tuned BERT-1seq Large Cased (s & 93.3 & 78.6 & 14.7 & 6 & 0 \\
	7 & InfoWord (large) & 93.1 & 72.7 & 20.4 & 12 & -5 \\
\end{longtable}
\end{center}

\begin{center}
    \footnotesize
    \rowcolors{2}{gray!15}{white}
    \begin{longtable}{llccccc}
    \caption{Model EM scores for against the \emph{adversarial} distribution
    shifts {\tt AddSent}.}\\
    \appendixtable{AddSent}{EM}
    	1 & XLNet (single model) & 89.9 & 71.2 & 18.7 & 2 & -1 \\
	2 & XLNET-123++ (single model) & 89.9 & 73.3 & 16.6 & 1 & 1 \\
	3 & XLNET-123 (single model) & 89.6 & 63.7 & 25.9 & 8 & -5 \\
	4 & SpanBERT (single model) & 88.8 & 66.3 & 22.5 & 5 & -1 \\
	5 & BERT+WWM+MT & 88.7 & 69.4 & 19.2 & 3 & 2 \\
	6 & Tuned BERT-1seq Large Cased (s & 87.5 & 65.8 & 21.7 & 6 & 0 \\
	7 & InfoWord (large) & 87.3 & 59.6 & 27.7 & 10 & -3 \\
\end{longtable}
\end{center}

\begin{center}
    \footnotesize
    \rowcolors{2}{gray!15}{white}
    \begin{longtable}{llccccc}
    \caption{Model EM scores for against the \emph{adversarial} distribution
    shifts {\tt AddOneSent}.}\\
    \appendixtable{AddOneSent}{EM}
    	1 & XLNet (single model) & 89.9 & 78.2 & 11.7 & 2 & -1 \\
	2 & XLNET-123++ (single model) & 89.9 & 80.3 & 9.6 & 1 & 1 \\
	3 & XLNET-123 (single model) & 89.6 & 73.5 & 16.1 & 6 & -3 \\
	4 & SpanBERT (single model) & 88.8 & 74.3 & 14.5 & 5 & -1 \\
	5 & BERT+WWM+MT & 88.7 & 75.6 & 13.1 & 4 & 1 \\
	6 & Tuned BERT-1seq Large Cased (s & 87.5 & 73.1 & 14.4 & 7 & -1 \\
	7 & InfoWord (large) & 87.3 & 67.5 & 19.8 & 11 & -4 \\
\end{longtable}
\end{center}

\FloatBarrier

\end{document}